\pdfoutput=1
\documentclass[11pt]{article}

\usepackage[preprint]{acl}

\usepackage{times}
\usepackage{latexsym}

\usepackage[T1]{fontenc}
\usepackage[utf8]{inputenc}

\usepackage{microtype}

\usepackage{inconsolata}

\usepackage{graphicx}

\usepackage{pifont}
\usepackage{array}
\newcommand*{\mline}[1]{%
\begingroup
    \renewcommand*{\arraystretch}{1.1}%
   \begin{tabular}[c]{@{}>{\raggedright\arraybackslash}c@{}}#1\end{tabular}%
  \endgroup
}

\newcommand{\myparagraph}[1]{\textbf{#1~~}}

\title{Emotion Identification for French in Written Texts: Considering their Modes of Expression as a Step Towards Text Complexity Analysis}

\author{Aline \'Etienne \\ ~~~~~~~~~~~~~~~~~~~~~~~~~~~~~~~~~~~~~~~~~~~~~~~~~~~~~~Univ. Paris-Nanterre, CNRS, MoDyCo -- Nanterre, France \\  \texttt{\normalsize acm.etienne@gmail.com}
        \And  Delphine Battistelli \\  \\  \texttt{\normalsize del.battistelli@gmail.com} \And
        Gw\'enol\'e Lecorv\'e \\ Orange -- Lannion, France\\ \texttt{\normalsize gwenole.lecorve@orange.com}}
\begin{document}
\maketitle
\begin{abstract}
The objective of this paper is to predict (A) whether a sentence in a written text expresses an emotion, (B) the mode(s) in which it is expressed, (C) whether it is basic or complex, and (D) its emotional category.
One of our major contributions, through a dataset and a model\footnote{\label{fn:huggingface}\url{https://huggingface.co/TextToKids}}, is to integrate the fact that an emotion can be expressed in different modes: from a direct mode, essentially lexicalized, to a more indirect mode, where emotions will only be suggested, a mode that NLP approaches generally don't take into account. 
Another originality is that the scope is on written texts, as opposed usual work focusing on conversational (often multi-modal) data. In this context, modes of expression are seen as a factor towards the automatic analysis of complexity in texts.
Experiments on French texts show acceptable results compared to the human annotators' agreement, and outperforming results compared to using a large language model with in-context learning (i.e. no fine-tuning).
\end{abstract}

\section{Introduction}
In Natural Language Processing (NLP), emotion detection and classification are often addressed in the context of interactions or conversations (e.g.,~\citep{poria2019emotion}), with either spoken, written (chats, forums, tweets) or multimodal datasets (e.g.,~\citep{busso2008iemocap,poria2018meld,chen2018emotionlines}). The goal is usually to identify the emotions felt by speakers in dialogic situations.
On the contrary, the analysis of emotions in non-conversational texts, like journalistic and encyclopedic texts or novels, is less developed in NLP. It indeed implies a different goal, which is no longer to characterize the emotional state of speakers but rather of characters/people in these texts. As pointed out in psycholinguistics, emotions in these types of texts are used---with more or less control by the writer---to capture the reader's attention. They also help to create a connection between the described situations and, thus, are a key factor in understanding (e.g., for children in \citealp{davidson_childrens_2001}). However, it is crucial that these emotions themselves are identified and understood. This leads to the idea that emotions can be considered as a factor of complexity, at least relative complexity in the terminology of~\citet{ehret2023measuring}, meaning it takes into account the difficulty perceived by speakers in terms of language learning or understanding. A text will thus be all the more complex as it contains emotions considered complex by a given type of speaker. In the case of children, for example, it is known that certain emotional categories are not accessible in early ages, and that their mode of expression (direct \textit{vs.} indirect or implicit) also plays a role in accessing their meaning.

\begin{table*}[t!]
\begin{center}
        \footnotesize
        \begin{tabular}{@{}p{7cm}||@{~}c@{~}||@{}c@{}|@{}c@{}|@{}c@{}|@{}c@{}||@{}c@{}|@{}c@{}||@{}c@{}|@{}c@{}|@{}c@{}|@{}c@{}|@{}c@{}|@{}c@{}|@{}c@{}|@{}c@{}|@{}c@{}|@{}c@{}|@{}c@{}|@{}c@{}|@{}c@{}}
         &
          (A) &
          \multicolumn{4}{@{}c@{}||}{(B) Expr.} &
          \multicolumn{2}{@{}c@{}||}{(C)} &
          \multicolumn{12}{@{}c@{}|}{(D)}
        \\
         &
          Emo. &
          \multicolumn{4}{@{}c@{}||}{mode} &
          \multicolumn{2}{@{}c@{}||}{Type} &
          \multicolumn{12}{@{}c@{}|}{Emotional category}
        \\
        ~~~~~~~~~~~~~~~\textbf{Sentence} (+ surrounding sentences) &
          \rotatebox{90}{\textbf{emo.}} &
          \rotatebox{90}{\textbf{beha.}} &
          \rotatebox{90}{\textbf{lab.}} &
          \rotatebox{90}{\textbf{disp.}} &
          \rotatebox{90}{\textbf{sug.}} &
          \rotatebox{90}{\textbf{bas.}} &
          \rotatebox{90}{\textbf{comp.}} &
          \rotatebox{90}{\textbf{adm.}} &
          \rotatebox{90}{\textbf{other}} &
          \rotatebox{90}{\textbf{ang.}} &
          \rotatebox{90}{\textbf{guilt}} &
          \rotatebox{90}{\textbf{dis.}} &
          \rotatebox{90}{\textbf{emb.}} &
          \rotatebox{90}{\textbf{pride}} &
          \rotatebox{90}{\textbf{jeal.}} &
          \rotatebox{90}{\textbf{joy}} &
          \rotatebox{90}{\textbf{fear}} &
          \rotatebox{90}{\textbf{surp.}} &
          \rotatebox{90}{\textbf{sad.}}
          \\
          \hline
        \begin{tabular}[c]{@{}p{7cm}@{}}\scriptsize How does the coronavirus spread? \textbf{Especially through respiratory droplets expelled by an infected person.} Respiratory droplets are small droplets of saliva that are released into the air when we talk, cough, or sneeze.\end{tabular} &
          ~~~~~ &
          ~~~~~ &
          ~~~~~ &
          ~~~~~ &
          ~~~~~ &
          ~~~~~ &
          ~~~~~ &
          ~~~~~ &
          ~~~~~ &
          ~~~~~ &
          ~~~~~ &
          ~~~~~ &
          ~~~~~ &
          ~~~~~ &
          ~~~~~ &
          ~~~~~ &
          ~~~~~ &
          ~~~~~ &
          ~~~~~ \\
           \hline
        \begin{tabular}[c]{@{}p{7cm}@{}}\scriptsize It is mainly celebrated in the Anglo-Saxon world. \textbf{Traditionally, children wear scary costumes.} They dress up as often despised and feared creatures such as ghosts, vampires, or witches and go door-to-door in the neighborhood, asking for candies or pastries.\end{tabular} &
          \mline{\ding{51}} &
           &
          \mline{\ding{51}} &
           &
           &
          \mline{\ding{51}} &
           &
           &
           &
           &
           &
           &
           &
           &
           &
           &
          \mline{\ding{51}} &
           &
           \\
           \hline
        \begin{tabular}[c]{@{}p{7cm}@{}}\scriptsize --- He succumbed after ingesting his herbal tea and a toxic substance, presumably cyanide. \textbf{From there, it was only a small step for Angus's mother to accuse the king of murder as she rushed towards her brother.}\\\scriptsize --- The herbal tea\dots\end{tabular} &
          \mline{\ding{51}} &
          \mline{\ding{51}} &
           &
           &
          \mline{\ding{51}} &
          \mline{\ding{51}} &
           &
           &
           &
          \mline{\ding{51}} &
           &
           &
           &
           &
           &
           &
           &
           &
           \\
           \hline
        \begin{tabular}[c]{@{}p{7cm}@{}}\scriptsize This summer, Nolita had to eat a sausage for the first time in a long time because there was nothing else. \textbf{"I forced myself," she said.} "It disgusted me, and I felt guilty," she recounted.\end{tabular} &
          \mline{\ding{51}} &
           &
          \mline{\ding{51}} &
           &
           &
          \mline{\ding{51}} &
          \mline{\ding{51}} &
           &
           &
           &
          \mline{\ding{51}} &
          \mline{\ding{51}} &
           &
           &
           &
           &
           &
           &
           \\
           \hline
        \begin{tabular}[c]{@{}p{7cm}@{}}\scriptsize At the Rome Olympics, the historic event takes place during the marathon: Ethiopian Abebe Bikila becomes the first athlete from black Africa to become an Olympic champion. \textbf{What's more, he achieved this feat... barefoot!} He had indeed developed the habit of running barefoot back home in Ethiopia.\end{tabular} &
          \mline{\ding{51}} &
           &
           &
          \mline{\ding{51}} &
          \mline{\ding{51}} &
          \mline{\ding{51}} &
          \mline{\ding{51}} &
           &
           &
           &
           &
           &
           &
          \mline{\ding{51}} &
           &
          \mline{\ding{51}} &
           &
          \mline{\ding{51}} &
          
        \end{tabular}
% \begin{center}
% \includegraphics[scale=0.85]{figures/examples.eps} 
\caption{Examples (translated from French) of sentences in context and reference labels for Tasks A, B, C, and D.}
\label{tab:examples}
\end{center}
% \vspace{-5mm}
\end{table*}

From these reflections on the question of emotions as a factor of complexity, this paper is oriented towards a better consideration of the diversity of modes of expression of emotions. We present a model and dataset that introduces the notion of mode of expression in addition to the usual information on emotional categories (e.g., joy, fear, etc.). In practice, the model classifies emotions in texts through four tasks: (A) predict whether a sentence contains an emotion or not; (B) if yes, how it is expressed (the \textit{mode}); (C) whether it is a basic or complex emotion category; and (D) in which emotional category it falls. Examples of these tasks on written texts from our dataset are given in Table~\ref{tab:examples}.
The model is built from the CamemBERT model~\citep{martin_camembert_2020} and data relies on a psycho-linguistically motivated annotation schema including different types of sources$^1$.
Evaluation shows that the proposed model outperforms approaches based on expert resources, non-neural architectures (SVM and XGBoost), and in-context learning using GPT-3.5.
A complementary human evaluation shows that the prediction errors made by the proposed model are generally in the same proportions as those made by humans. Finally, the paper discusses interactions between expression modes and emotional categories.
While complexity analysis is the motivation of our work, the paper is restricted to Tasks A-D. Application to complexity analysis are left for future work.

% Plan announcement
Section~\ref{sec:literature} browses the literature on emotion identification in written texts, particularly in NLP. Sections~\ref{sec:tasks}, \ref{sec:data} and \ref{sec:model} detail the tasks addressed, the associated data, and the proposed model, respectively. Section~\ref{sec:eval} reports the experiments and results.

%%%%%%%%% LITERATURE REVIEW %%%%%%%%%
\section{Framework and Related Work}
\label{sec:literature}

This section provides a brief overview of the framework for emotion analysis in which the paper is situated and which justifies the choice of schema and data (annotated with this schema). It also positions our work among studies in NLP.

\subsection{The Analysis of Emotions as a Complexity Factor of a Text }\label{subsec:annotation}

In psycholinguistics, the key role of characters' emotions on text comprehension is well-documented (e.g., \citealp{dijkstra_character_1995,dyer_role_1983}). Among recent works, two influencing factors have been highlighted in children's understanding of emotions, and thus of the texts themselves: the \textit{type of emotion} expressed, basic or complex---the complex emotions (e.g., pride, shame) being more difficult to grasp as they require knowledge of social norms---\citep{davidson_role_2006,blanc_production_2017}; as well as \textit{the way emotions are expressed}~\citep{creissen-blanc2017}), directly \textit{via} an emotional label, indirectly through the mention of an emotional behavior, or through the description of an emotional situation, the latter being the most difficult to understand. Of course, the notion of emotional category is also addressed in psycholinguistics, and it has been shown that some categories take longer to be mastered by children (e.g., \citealp{baron-cohen_emotion_2010}).

On the NLP side, several works (see e.g.,~\citealp{bostan_analysis_2018,acheampong_text-based_2020,ohman_emotion_2020}) highlight the great heterogeneity of emotion annotation schemas---and annotated corpora---, thus clearly demonstrating the difficulty of modeling emotions and, in the end, of analyzing them. This heterogeneity ranges from the notions (e.g., the number and types of emotional categories) and the type of data studied (journals, tweets, etc.) up to the annotation procedures (\textit{crowdsourcing}, annotation by experts) and evaluation methods implemented (e.g., with or without agreement between annotators). Although some works strive to take into account broader sets of notions and linguistic cues to analyze emotions (e.g., \citealp{casel-etal-2021-emotion, kim_analysis_2019}), the most commonly used concept remains the notion of \textit{emotional category}, often approached through a list of basic emotions introduced either by \citet{ekman_argument_1992} (anger, disgust, fear, joy, sadness, and surprise) or \citet{plutchik_general_1980} (Ekman's categories, anticipation, and trust), with a focus on one way of expressing emotions: the emotional lexicon.
As highlighted in~\citep{klinger2023eventcentric} and in~\citep{Troiano_klinger_2023}, a few very recent approaches in NLP aim to acquire a deeper understanding of the textual units that support the evocation of emotions outside of directly emotional lexical terms (e.g., "happy", "anger"). These approaches are then inspired by psychological and/or linguistic models of emotions. We adopt the same approach here because we aim to capture both direct and indirect modes of expression of emotions in texts. Like~\citet{Troiano_klinger_2023}, we seek to assess to what extent computational models can capture emotions expressed indirectly (e.g., via the description of situations that are/ associated with emotions with regard to social norms and conventions). More specifically, our work adopts the framework proposed by \citet{etienne_psycho-linguistically_2022}, which proposes a detailed annotation schema of emotions for French.
To our knowledge, this is the only work with the explicit objective of analyzing emotions in texts by addressing both direct and indirect modes of expression.

\subsection{Automatic Identification of Emotions}
\label{subsec:systems}
In NLP, the analysis of emotions in texts is generally treated as a classification task. The previously mentioned heterogeneity of annotation schemas and annotated corpora is then reflected in the diversity of predicted classes, the granularity of elements to be classified, and the methods for developing and evaluating classifiers. The way results are presented thus also varies from one paper to another, making performance comparison more difficult.

The focus is often on the classification of basic emotions~\citep{strapparava_semeval-2007_2007,mohammad_emotional_2012,abdaoui_feel_2017,demszky_goemotions_2020,ohman_xed_2020,bianchi_feel-it_2021}, although some works use a mix of basic and complex emotions~\citep{balahur_detecting_2012,fraisse_utiliser_2015,abdaoui_feel_2017,mohammad_semeval-2018_2018,liu_dens_2019,demszky_goemotions_2020}.
Moreover, there is a long history of building and using emotional lexicons, and the diversity of linguistic markers of emotions is not systematically taken into account, although it is mentioned in several works~\citep{alm_emotions_2005,mohammad_emotional_2012,kim_who_2018,demszky_goemotions_2020}). Some works nevertheless study other means of expression. For example, \citet{kim_analysis_2019} analyze non-verbal expressions of emotions by characters in a corpus of fanfictions (e.g., looks, gestures). \citet{balahur_detecting_2012} aim to detect indirect emotions. These works have the limitation of focusing each time only on one mode of expression, thus leaving aside the complementarities between modes. For their part, based on Scherer's model of emotional components process \citeyearpar{scherer_what_2005}, \citet{casel-etal-2021-emotion} annotated and then predicted several components of emotions, such as physiological symptoms and motor expressions of emotions, or the cognitive evaluation of events. Although \citep{casel-etal-2021-emotion} deal with a broader set of cues, these are not rigorously motivated linguistically. Therefore, relying on~\citep{etienne_psycho-linguistically_2022}, the originality of our work lies in taking into account different modes of expression of emotions.

% Types of models, inputs, etc.
Historically, \textit{Support Vector Machine} (SVM) models have been widely used to classify sentences~\citep{aman_identifying_2007,mohammad_emotional_2012}) or texts~\citep{abdaoui_feel_2017,balahur_detecting_2012,fraisse_utiliser_2015,mohammad_emotional_2012} according to the emotional category they express. Until the advent of embeddings, the inputs were mainly symbolic: bags of words or \textit{n}-grams, features based on emotional resources such as WordNetAffect~\citep{aman_identifying_2007,balahur_detecting_2012, strapparava_semeval-2007_2007} or emotional lexicons~\citep{strapparava_semeval-2007_2007,abdaoui_feel_2017,kim_who_2018}.
Today, neural networks~\citep{kim_who_2018} and Transformer architectures~\citep{liu_dens_2019,demszky_goemotions_2020,ohman_xed_2020,bianchi_feel-it_2021} obviously dominate the state of the art. As for French, no Transformer model has yet been proposed to our knowledge.

\section{Tasks}
\label{sec:tasks}

%Built in the global perspective of analyzing text complexity,
Built in the global perspective of enabling the analysis of emotions as a complexity factor, our work thus takes into account two key elements to address the complexity of an emotion: its category and its mode of expression. The goal is to propose a Transformer model for 4 classification tasks (noted A, B, C, and D) at the \textit{sentence} level, as opposed to the text level (this can, for example, allow studying how the presence of emotions evolves along a text). Sentences can contain several emotions, as in Table~\ref{tab:examples}. Classifications are therefore multi-label.

\myparagraph{Task A: Presence of Emotion}
\label{subsec:emo}
The first task aims to predict the presence of emotional information in a given sentence (binary prediction).
%It reflects a first factor of textual complexity. Indeed, it has been demonstrated that the mere presence of emotional information (regardless of the emotional category expressed or how it is expressed) can improve text comprehension (e.g., for children in \citealp{davidson_childrens_2001}).

\myparagraph{Task B: Mode of Expression}
\label{subsec:mode}
The mode of expression focuses on the linguistic means used to convey the presence of an emotion in a text.
%It allows for a finer and broader linguistic analysis of emotion, and it also reflects a marker of text complexity.
Following \citet{etienne_psycho-linguistically_2022}, 4 modes are considered:
the \myparagraph{labeled emotions} directly indicated by a term from the emotional lexicon (e.g., \textit{happy}, \textit{scared}); the
\myparagraph{behavioral emotions} which rely on the description of an emotional behavior, such as physiological manifestations (e.g., \textit{crying}, \textit{smiling}) or other behaviors (e.g., \textit{slapping someone});
the \myparagraph{displayed emotions} which are expressed by very heterogeneous surface linguistic features of statements that mainly reflect the emotional state of the writer (e.g., interjections, short sentences);
the \myparagraph{suggested emotions} which emanate from the description of a situation generally associated with an emotional feeling according to social norms and conventions (e.g., \textit{seeing a good friend after a long period} suggests joy).

\myparagraph{Task C: Type of Emotion}
\label{subsec:type}
Task C aims to predict the presence of \textit{basic} and \textit{complex} emotion types (2 simultaneous binary predictions).
To our knowledge, this notion has not yet been studied as such in automatic emotion analysis (although the emotional categories \textit{basic} and \textit{complex} have been used in NLP (\textit{cf.} section~\ref{subsec:systems})). This is probably due to the fact that the type of an expressed emotion is directly related to its emotional category. However, the type of emotion is in itself a marker of complexity, as we have seen.
%since complex emotions are more difficult to understand for children, hence our specific interest.

\myparagraph{Task D: Emotional Category}
\label{subsec:cat}
In accordance with \citet{etienne_psycho-linguistically_2022}, Task~D is designed to label 11+1 emotional categories, namely the 6 basic emotions of Ekman (\textit{anger}, \textit{disgust}, \textit{fear}, \textit{joy}, \textit{sadness}, and \textit{surprise}) and 5 complex emotions (\textit{admiration}, \textit{embarrassment}, \textit{guilt}, \textit{jealousy}, and \textit{pride}). A last category, named \textit{other}, is used to capture markers that express any other emotion (e.g., hate, contempt, love, \textit{etc.}).

\section{Data}
\label{sec:data}
%
% General description of the corpus we use

\begin{table}[t!]
\begin{center}
\includegraphics[scale=0.8]{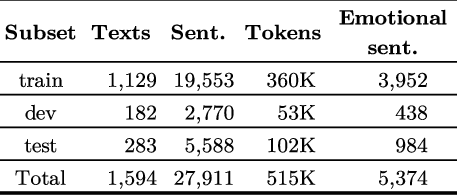} 
\caption{Statistics over the dataset.}
\label{tab:stats_datasets}
\end{center}
\vspace{-5mm}
\end{table}

As detailed in Table~\ref{tab:stats_datasets}, our proposed corpus consists of 1,594 French texts (28K sentences, 515K words) intended for children aged 6 to 14 years, divided into 3 types: mainly journalistic texts (91\% of the sentences), encyclopedic articles (9\%), and novels (1\%).
Annotations conducted by 6~experts associate emotional units (segments) in the texts with their mode of expression and emotional category, following the annotation schema and guide in~\citep{etienne_psycho-linguistically_2022}. Inter-annotator agreements are presented in Appendix~\ref{app:iaa}.
These annotations are then merged from the segment level to the sentence level. Thus, a given sentence may cover several emotional units. The presence of emotions and the types of emotions have been derived from the mode of expression and emotional category labels. In the end, each sentence is associated with a vector of 19 booleans (again, see Table~\ref{tab:examples}).
The data are divided into training, development, and test sets (70/10/20\% of the sentences, respectively), such that all sentences from a text are in the same subset, in order to avoid a training bias on the peculiarities of the texts (e.g., the name of a character).

Table~\ref{tab:prop_labels} presents the proportion of labels within the corpus. Overall, the proportions are comparable from one subset to another. Several imbalances appear within the tasks.
%
% Presence of emotion
\myparagraph{(A)} Only 15-20\% of sentences are emotional.
%
% Modes of expression
\myparagraph{(B)} Modes of expression are quite evenly distributed, 'labeled' being the least frequent (3\% of sentences) and \textit{suggested} the most common (6\%). The sums of the percentages of each mode are higher than the percentages of the \textit{emotional} label because a sentence can certain emotions are conveyed by several modes and a sentence can also contain several emotional units whose respective modes differ.
%
% Types of emotion
\myparagraph{(C)} The labels of emotion types are very unbalanced, with a clear dominance of basic emotions. The emotional category 'other' (task D) is not associated with any type of emotion, hence the fact that the sum of the percentages \textit{basic} and \textit{complex} is lower than that of emotional sentences.
%
% Emotional categories
\myparagraph{(D)} The labels of emotional categories are unbalanced, with percentages always below 5\% of sentences. The categories \textit{anger}, \textit{fear}, \textit{joy}, \textit{sadness}, \textit{surprise}, and \textit{other} are dominant, while others are very rare (\textit{disgust}, \textit{guilt}, and \textit{jealousy}).

\begin{table}[t!]
\begin{center}
\includegraphics[scale=0.8]{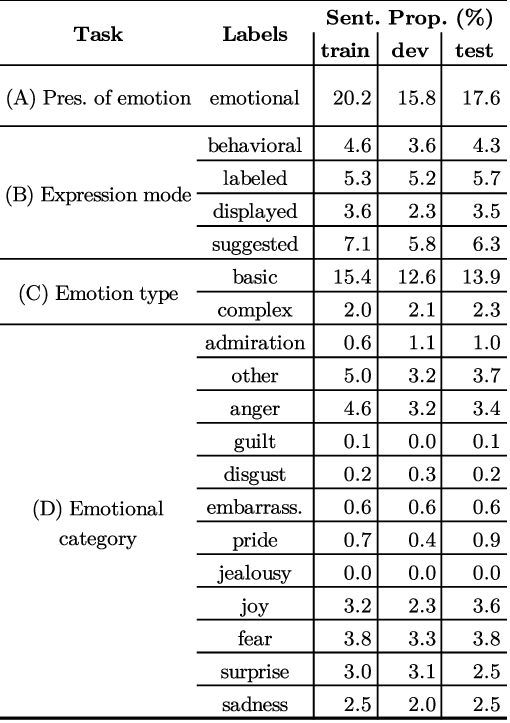} 
\caption{Distribution of labels}
\label{tab:prop_labels}
\end{center}
\end{table}

\section{Proposed Model}
\label{sec:model}
All tasks are learned together, leading to a single proposed model. This model results from fine-tuning the base version of the pre-trained CamemBERT model~\citep{martin_camembert_2020}. It is a BERT-type encoder model with 110 million parameters and 12 BERT layers. It was pre-trained on 138GB of French texts~\cite{suarez2019asynchronous}.
Although more recent and larger generative language models like Llama2 or Mistral would likely yield better results, the choice of a reasonably sized model is motivated by two reasons. Firstly, our goal is to demonstrate that, unlike several other tasks in NLP, fine-grained emotion characterization in texts cannot be achieved by leveraging large generic (i.e., non-specialized) language models via in-context learning (i.e., without fine-tuning). Secondly, our work aims for a lightweight solution, so that emotion characterization can be integrated as a processor for analyzing text complexity in a massive collection of texts from a public search engine. Thus, while fine-tuning larger models is part of our future work, this paper does not address it.

We fine-tune the CamemBERT model by replacing its last token prediction layer with a binary classification layer of the size of the number of labels, using binary cross-entropy as the loss function. The fine-tuning involves all model weights, i.e., no layers are frozen.
Following prototyping work on the development set, the final model is not directly learned from CamemBERT. An initial fine-tuning is conducted on Task A alone for 3 epochs (classification layer of size $1$), then the final model is fine-tuned on all tasks starting from this intermediate model for an additional 6 epochs (the final classification layer is replaced by a fresh layer of size $19$). The optimizer is Adam with a learning rate of $10^{-5}$ (no decay) and batches of 8~examples.

%%%%%%%%% DEVELOPMENT WORKS %%%%%%%%%
\label{sec:dev_exp}
%
% Class distribution: many imbalances
Other experiments were conducted on the development set, for example, on the choice of a window around sentences, class weighting or not, or the choice of initial fine-tuning only on Task A or not. Ultimately, the results presented are those of the best strategy obtained on the development set averaging results over 3 training runs with different random initializations. Notably, a weighting between classes is adopted so as not to overly favor the majority classes. The maximum weighting factor is capped at~50 to, conversely, not give too much importance to very rare classes. Finally, the model takes as input a triplet of sentences where the target sentence to be labeled is surrounded by its preceding and following sentence in the form \texttt{before: \{previous\}</s>current: \{target\}</s>after: \{next\}</s>}. Details of these developmental work can be found in~\cite{etienne2023analyse}.

%%%%%%%%% AUTOMATIC AND HUMAN EVALUATION %%%%%%%%%
\section{Automatic and Human Evaluations}
\label{sec:eval}

\subsection{Comparison with Other Models}
\label{subsec:baselines}

\begin{table}[t!]
\begin{center}
\includegraphics[scale=0.8]{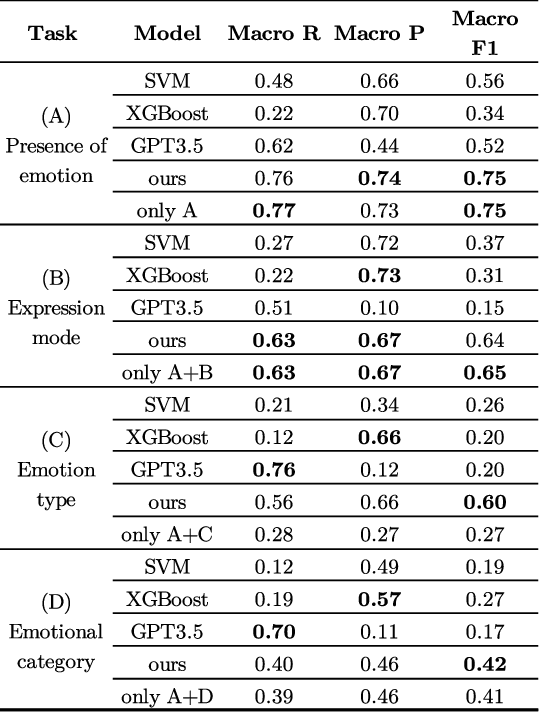} 
\caption{Model performances (averages over 3 runs, all standard deviations are below $0.02$).}
\label{tab:perf_baselines_and_E}
\end{center}
\end{table}

The proposed model is compared to three other types of models.
%
% SVM model
\textbf{SVM} models were trained as they are a historical approach in the field.
Two types of input features were used: (i) bag-of-tokens where tokens come from the CamemBERT tokenizer, restricted to those from the training set, resulting in input vectors of dimension 18,437; (ii) sentence embeddings of size 768 obtained with SentenceTransformer~\cite{reimers2019sentence} and CamemBERT\footnote{\url{https://huggingface.co/dangvantuan/sentence-camembert-base}}.
% 
% XGBoost model
\textbf{XGBoost} models were trained as it is a more recent, lightweight, and competitive technique for many classification tasks, especially with unbalanced data~\cite{chen2016xgboost}. The input features are the same as for the SVMs.
% 
% GPT-3.5
Our approach is compared to \textbf{GPT-3.5}~\cite{ouyang2022training}. For a given input sample, GPT-3.5 is incrementally solicited to annotate it with binary labels (yes/no). Consecutively for each task and label, a natural language description of what is expected is provided to the model before asking for a response, accompanied by examples from the training set for each label. Different prompts were tested (see details in Appendix~\ref{app:gpt35}). The chosen one reports 2 to 4 positive examples per label. Unlike SVM, XGBoost, and our model, this approach is not economical but it does require any training.

Table~\ref{tab:perf_baselines_and_E} summarizes the performances on the test set of the best models for each task---for SVM and XGBoost, the bag-of-tokens feature; for GPT-3.5, the prompts without negative examples---and compares them to our model.
Models are evaluated through recall (R), precision (P), and F1 scores.
Overall, it appears that our proposed model significantly outperforms SVMs, XGBoost, and GPT-3.5 in terms of F1 scores for all tasks, with values almost double those of the best-ranked model for tasks B, C, and D. It seems especially that all other models tend to favor either recall (GPT-3.5) or precision (SVM, XGBoost), while our model is balanced.
Finally, the poor results of GPT-3.5 show that the task is difficult and requires fine-tuning.

Table~\ref{tab:perf_baselines_and_E} also reports the comparison with 4~variants of the model where Tasks~B, C, and~D are trained separately on top of~A. These results show that multi-task does degrade the performance but it does not really improve them neither, except for Task~C where guessing the category probably helps. This may lead one to consider that the interaction between mode and category is not very strong. Further discussions are exposed in Section~\ref{sec:mode_category}.

\subsection{Comparison with Related Work}
\label{subsec:comparison}

In the absence of truly similar work to ours, this section reports additional results to give a better intuition of the performance of our model.

\begin{table}[!t]
\begin{center}
\includegraphics[scale=0.8]{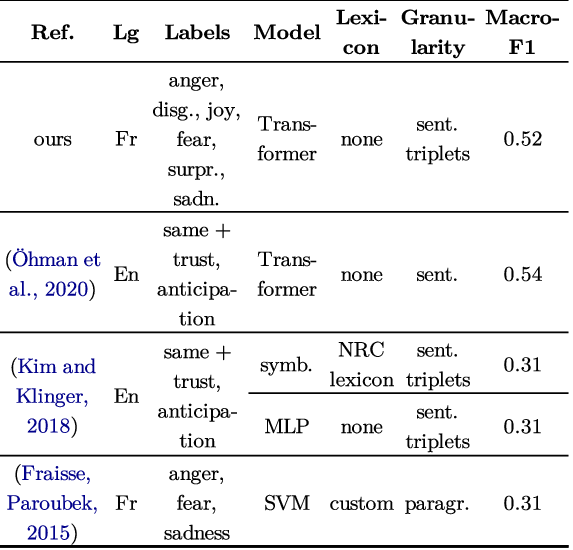}
\caption{Comparison elements with close works.}
\label{tab:perf_closest}
\vspace{-1mm}
\end{center}
\end{table}

\myparagraph{Closest Comparable Works}
Table~\ref{tab:perf_closest} summarizes the performances of the three closest works we could find in the literature. They were chosen because they all predict labels at a granularity close to that of the sentence. \citep{ohman_xed_2020} allows a comparison with another Transformer model; \citep{fraisse_utiliser_2015} with another work in French; and \citep{kim_who_2018} with a method that works at the level of linguistic markers (as opposed to the phrasal or textual level). All focus solely on emotional categories.
The results show that our model is competitive.

\begin{table}[!t]
\begin{center}
\includegraphics[scale=0.8]{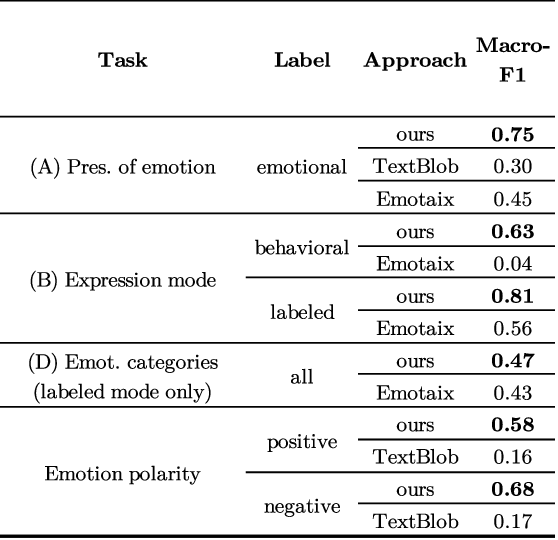} 
\caption{Comparison with tools available for French.}
\label{tab:perf_tools}
\vspace{-5mm}
\end{center}
\end{table}

\myparagraph{Implementations Based on Existing Resources}

In the absence of dedicated model for French, two resources are currently available in French if one wants to consider emotion identification in texts: TextBlob ({\small\url{https://textblob.readthedocs.io/}}), a sentiment analysis library that integrates a French lexicon where terms are associated with a negative and positive weight reflecting their polarity; Emotaix~\cite{piolat2009example}, another lexicon comprising associations (i) of terms with emotional categories for the labeled mode only, and (ii) other terms with the behavioral mode (but this time without information on the emotional category).
Several tasks managed by our model were replicated \textit{via} TextBlob and Emotaix. To account for the differences between these resources and our proposed model, Task B was limited to only the behavioral and labeled modes and Task D to the labeled mode.
Moreover, our model was tested on a task of predicting emotional polarity on our test set since TextBlob is designed for this use. To predict polarity \textit{via} our model, categories were predicted and empirically projected towards positive or negative polarity (e.g., \textit{anger} is \textit{negative}, \textit{joy} is \textit{positive}).
As shown in Table~\ref{tab:perf_tools}, our model performs significantly better than TextBlob and Emotaix, including in the emotional polarity task for which it was not specifically designed. The only task where the competition remains is the prediction of categories when the mode is labeled, which is the easiest situation compared to considering all modes.

%%%%%%%%% HUMAN EVALUATION %%%%%%%%%
% \myparagraph{Human Evaluation}
\subsection{Human Evaluation}
\label{sec:human_eval}
Given the difficulty of the tasks considered, it is appropriate to cross-reference the automatic evaluation with a human analysis, particularly to give an intuition of what the observed prediction errors represent.
A perceptual validation experiment was thus conducted with three experts in text complexity and emotions. Each of them was informed of the tasks and the definitions of labels in psycho-linguistics and linguistics. They were then each confronted with 150 sentences from the test set and their labels for emotional category and mode of expression. These labels came either from the human reference annotations or from the predictions of our model. For each label, the experts had to say whether they agreed or not with the proposed annotation. Of course, they were not aware of the origin of the labels.

\begin{table}[t!]
\begin{center}
\includegraphics[scale=0.8]{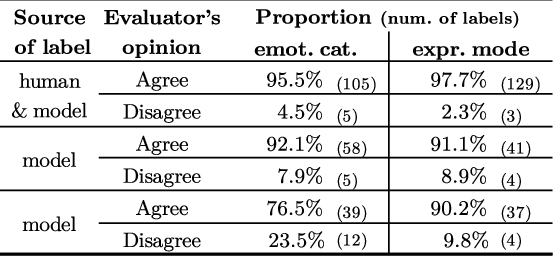} 
\caption{Experts' agreement regarding predictions common to the human annotator and our model, or specific to each.}
\vspace{-5mm}
\label{tab:accord_experts}
\end{center}
\end{table}

% \footnote{We considered that the experts agree with a label when at least 2 out of 3 declared that they agreed with the label.}
Table~\ref{tab:accord_experts} reports the experts' agreement rates with the proposed labels, depending on the source of the label.
Although the strongest agreement is when the human and model labels match (\textit{human \& model}), the agreement scores are generally very high, especially for the mode of expression.
These results thus tend to show that, even when the model predicts differently from the reference, the prediction is generally considered relevant by human experts. This demonstrates that our model is able to generalize correctly and that the F1 scores from previous experiments underestimate the perceived quality of the model's predictions.

% \subsection{Detailed Analysis}
% \label{subsec:analysis}

% This section details the results for each task, then presents the human evaluation.

% \myparagraph{Results by Label}
\subsection{Results by Label}

Table~\ref{tab:perf_modele_E} presents the results of our classifier on all labels of all tasks from the test set. Additional observations can be made as follows.
%
% Expression modes
Regarding expression modes (B), labeled emotions are very well recognized (F1 > 0.8), unlike suggested emotions (F1 < 0.5). This is not surprising as labeled emotions are the easiest to identify for a human annotator, while suggested emotions have the largest part of interpretation.
% Types of emotion
Performance for emotion types (C) seems in turn linked to the results on emotional categories, since the \textit{basic} label is, as intuition would suggest, better recognized than the \textit{complex} label.
%
% Emotional categories
Finally, regarding emotional categories (D), three of them are never predicted (\textit{guilt}, \textit{disgust}, and \textit{jealousy}). These are the rarest labels in the training set, probably too rare for the model to learn to predict them.
Indeed, the best-predicted emotional categories are the basic emotions, more frequent, namely the labels \textit{surprise}, \textit{fear}, and \textit{anger} (see Table~\ref{tab:prop_labels}). However, while \textit{surprise} is the best-predicted label of Task D, it is not the most represented in the training set. Conversely, \textit{sadness} is not well recognized, even though it is one of the most frequent emotional categories. From our additional analyses, this seems to be explained by sometimes strong interactions between the notions of expression mode and emotional category.
% Indeed, if there is a strong association between an emotional category and an expression mode in the training set, the model will better recognize the category when it is expressed by this mode.

% Results of the best model for Task E
\begin{table}[t!]
\begin{center}
\includegraphics[scale=0.8]{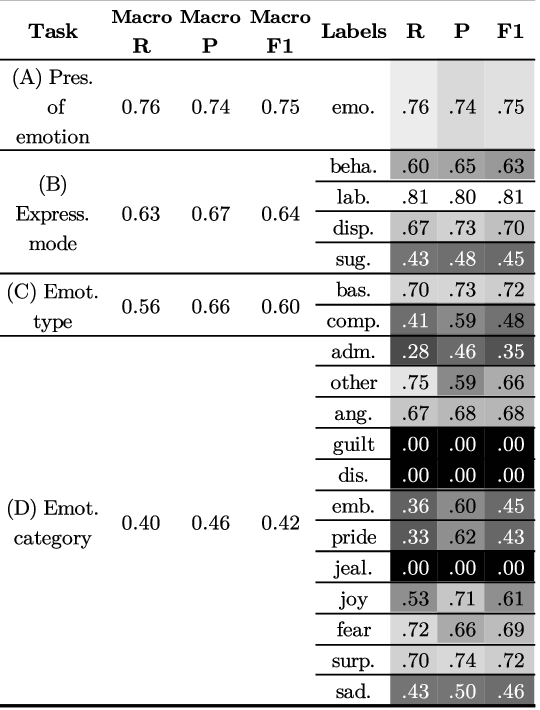} 
\caption{Detailed performances of our model}
\label{tab:perf_modele_E}
\end{center}
\end{table}

\subsection{Correlation Between Mode and Category}
\label{sec:mode_category}

\begin{table}[t!]

\begin{center}
\includegraphics[scale=0.8]{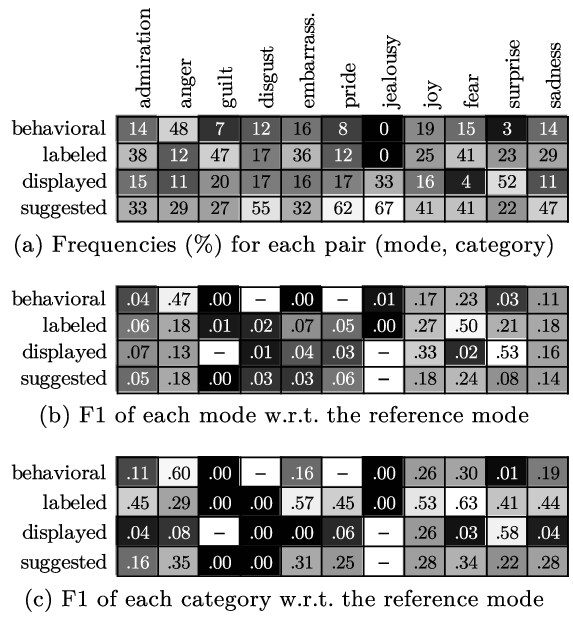} 
\caption{Interactions between mode and category: frequency of cooccurence in the reference (a) ; impact of category and mode on the prediction of the other (b and~c).}
\vspace{-2mm}
\label{tab:mode_category}
\end{center}
\end{table}

This final section assesses how specific modes impact the prediction of emotional categories, and \textit{vice-versa}. As background information, {Table~\ref{tab:mode_category}.a} reports the cooccurrence relative frequencies between modes and categories in the test set.

Table~\ref{tab:mode_category}.b explores how the mode's predictability varies with the emotional category expressed. Strong associations between emotional categories and expression modes enhance mode recognition, but the suggested mode remains challenging across categories, highlighting its complexity for both our model and human annotators.

Then, Table~\ref{tab:mode_category}.c indicates that F1 scores are generally higher when emotions are expressed through the labeled mode. However, \textit{anger} and \textit{surprise} deviate from this pattern, performing better in \textit{behavioral} and \textit{displayed} modes, respectively.
The effectiveness of our in recognizing emotions like behavioral \textit{anger} and displayed \textit{surprise} is influenced by their strong association with these modes in the training data. However, factors such as the rarity of the emotion in the training set (e.g., \textit{disgust}) and the linguistic characteristics of the mode also play significant roles. For instance, \textit{joy} and \textit{fear} are better recognized when labeled, despite being frequently suggested, due to the inherent challenge in recognizing the suggested mode. 
% For example, the category \textit{surprise}, which is mainly \textit{displayed} in the corpus, is on average 14 times better recognized when expressed by this mode compared to other modes. Similarly, \textit{anger}, mainly \textit{behavioral} in the corpus, is 4 times better predicted in this mode.

%%%%%%%%% CONCLUSION AND PERSPECTIVES %%%%%%%%%
\section{Conclusion and Perspectives}

In this paper, we have addressed the task of detecting and classifying emotions in written texts, as opposed to conversational data. Due to the applicative perspective towards complexity analysis, we introduced a dataset of French texts and a model which, additionnally to the usual notion of emotion categories, takes into account their direct but also indirect modes of expression.
The experiments show that this model performs well compared to other approaches, comparable works, and solutions from off-the-shelf resources. Human evaluation has shown that this level is almost equivalent to what humans can do.

Future work includes intra-sentential predictions, allowing the inclusion of other notions, such as that of experiencer, fine-tuning of larger models to propose a leaderboard on our dataset, and applications to text complexity analysis.

\bibliography{submission}

\begin{thebibliography}{43}
\expandafter\ifx\csname natexlab\endcsname\relax\def\natexlab#1{#1}\fi

\bibitem[{Abdaoui et~al.(2017)Abdaoui, Azé, Bringay, and
  Poncelet}]{abdaoui_feel_2017}
Amine Abdaoui, Jérôme Azé, Sandra Bringay, and Pascal Poncelet. 2017.
\newblock Feel: a french expanded emotion lexicon.
\newblock \emph{Language Resources and Evaluation}, 51(3):833--855.
\newblock Publisher: Springer.

\bibitem[{Acheampong et~al.(2020)Acheampong, Wenyu, and
  Nunoo-Mensah}]{acheampong_text-based_2020}
Francisca~Adoma Acheampong, Chen Wenyu, and Henry Nunoo-Mensah. 2020.
\newblock Text-based emotion detection: {Advances}, challenges, and
  opportunities.
\newblock \emph{Engineering Reports}, 2(7):e12189.
\newblock Publisher: Wiley Online Library.

\bibitem[{Alm et~al.(2005)Alm, Roth, and Sproat}]{alm_emotions_2005}
Cecilia~Ovesdotter Alm, Dan Roth, and Richard Sproat. 2005.
\newblock \href {https://aclanthology.org/H05-1073} {Emotions from {Text}:
  {Machine} {Learning} for {Text}-based {Emotion} {Prediction}}.
\newblock In \emph{Proceedings of {the Human} {Language} {Technology}
  {Conference} and {Conference} on {Empirical} {Methods} in {Natural}
  {Language} {Processing} (HLT-EMNLP)}, pages 579--586, Vancouver, British
  Columbia, Canada. Association for Computational Linguistics.

\bibitem[{Aman and Szpakowicz(2007)}]{aman_identifying_2007}
Saima Aman and Stan Szpakowicz. 2007.
\newblock Identifying expressions of emotion in text.
\newblock In \emph{Proceedings of the International {Conference} on {Text},
  {Speech} and {Dialogue} (TSD)}, pages 196--205. Springer.

\bibitem[{Balahur et~al.(2012)Balahur, Hermida, and
  Montoyo}]{balahur_detecting_2012}
Alexandra Balahur, Jesús~M Hermida, and Andrés Montoyo. 2012.
\newblock Detecting implicit expressions of emotion in text: {A} comparative
  analysis.
\newblock \emph{Decision support systems}, 53(4):742--753.
\newblock Publisher: Elsevier.

\bibitem[{Baron-Cohen et~al.(2010)Baron-Cohen, Golan, Wheelwright, and
  Granader}]{baron-cohen_emotion_2010}
Simon Baron-Cohen, Ofer Golan, Sally Wheelwright, and Yael Granader. 2010.
\newblock \href {https://doi.org/10.3389/fnevo.2010.00109} {Emotion {Word}
  {Comprehension} from 4 to 16 {Years} {Old}: {A} {Developmental} {Survey}}.
\newblock \emph{Frontiers in Evolutionary Neuroscience}, 0.
\newblock Publisher: Frontiers.

\bibitem[{Bianchi et~al.(2021)Bianchi, Nozza, and Hovy}]{bianchi_feel-it_2021}
Federico Bianchi, Debora Nozza, and Dirk Hovy. 2021.
\newblock Feel-it: {Emotion} and sentiment classification for the italian
  language.
\newblock In \emph{Proceedings of the {Workshop} on {Computational}
  {Approaches} to {Subjectivity}, {Sentiment} and {Social} {Media} {Analysis}},
  pages 76--83.

\bibitem[{Blanc and Quenette(2017)}]{blanc_production_2017}
Nathalie Blanc and Guy Quenette. 2017.
\newblock \href {https://doi.org/10.3917/enf1.174.0503} {La production
  d'inférences émotionnelles entre 8 et 10 ans : quelle méthodologie pour
  quels résultats ?}
\newblock \emph{Enfance}, 4(4):503--511.
\newblock Place: Paris Publisher: NecPlus.

\bibitem[{Bostan and Klinger(2018)}]{bostan_analysis_2018}
Laura-Ana-Maria Bostan and Roman Klinger. 2018.
\newblock \href {https://aclanthology.org/C18-1179} {An {Analysis} of
  {Annotated} {Corpora} for {Emotion} {Classification} in {Text}}.
\newblock In \emph{Proceedings of the {International} {Conference} on
  {Computational} {Linguistics} (COLING)}, pages 2104--2119, Santa Fe, New
  Mexico, USA. Association for Computational Linguistics.

\bibitem[{Busso et~al.(2008)Busso, Bulut, Lee, Kazemzadeh, Mower, Kim, Chang,
  Lee, and Narayanan}]{busso2008iemocap}
Carlos Busso, Murtaza Bulut, Chi-Chun Lee, Abe Kazemzadeh, Emily Mower, Samuel
  Kim, Jeannette~N Chang, Sungbok Lee, and Shrikanth~S Narayanan. 2008.
\newblock Iemocap: Interactive emotional dyadic motion capture database.
\newblock \emph{Language resources and evaluation}, 42:335--359.

\bibitem[{Casel et~al.(2021)Casel, Heindl, and
  Klinger}]{casel-etal-2021-emotion}
Felix Casel, Amelie Heindl, and Roman Klinger. 2021.
\newblock \href {https://aclanthology.org/2021.konvens-1.5} {Emotion
  recognition under consideration of the emotion component process model}.
\newblock In \emph{Proceedings of the Conference on Natural Language
  Processing}, pages 49--61, D{\"u}sseldorf, Germany. KONVENS 2021 Organizers.

\bibitem[{Chen et~al.(2018)Chen, Hsu, Kuo, Ku et~al.}]{chen2018emotionlines}
Sheng-Yeh Chen, Chao-Chun Hsu, Chuan-Chun Kuo, Lun-Wei Ku, et~al. 2018.
\newblock Emotionlines: An emotion corpus of multi-party conversations.
\newblock \emph{arXiv preprint arXiv:1802.08379}.

\bibitem[{Chen and Guestrin(2016)}]{chen2016xgboost}
Tianqi Chen and Carlos Guestrin. 2016.
\newblock Xgboost: A scalable tree boosting system.
\newblock In \emph{Proceedings of the ACM SIGKDD international conference on
  knowledge discovery and data mining}, pages 785--794.

\bibitem[{Creissen and Blanc(2017)}]{creissen-blanc2017}
S.~Creissen and N.~Blanc. 2017.
\newblock \href {https://doi.org/https://doi.org/10.1016/j.psfr.2015.07.006}
  {Quelle représentation des différentes facettes de la dimension
  émotionnelle d’une histoire entre l’âge de 6 et 10ans ? apports
  d’une étude multimédia}.
\newblock \emph{Psychologie Française}, 62(3):263--277.
\newblock Cognition et multimédia : les atouts du numérique en situation
  d’apprentissage.

\bibitem[{Davidson(2006)}]{davidson_role_2006}
Denise Davidson. 2006.
\newblock \href {https://doi.org/10.1007/s11031-006-9037-6} {The {Role} of
  {Basic}, {Self}-{Conscious} and {Self}-{Conscious} {Evaluative} {Emotions} in
  {Children}’s {Memory} and {Understanding} of {Emotion}}.
\newblock \emph{Motivation and Emotion}, 30(3):232--242.

\bibitem[{Davidson et~al.(2001)Davidson, Luo, and
  Burden}]{davidson_childrens_2001}
Denise Davidson, Zupei Luo, and Matthew~J. Burden. 2001.
\newblock \href {https://doi.org/10.1080/0269993004200105} {Children's recall
  of emotional behaviours, emotional labels, and nonemotional behaviours:
  {Does} emotion enhance memory?}
\newblock \emph{Cognition and Emotion}, 15(1):1--26.
\newblock Place: United Kingdom Publisher: Taylor \& Francis.

\bibitem[{Demszky et~al.(2020)Demszky, Movshovitz-Attias, Ko, Cowen, Nemade,
  and Ravi}]{demszky_goemotions_2020}
Dorottya Demszky, Dana Movshovitz-Attias, Jeongwoo Ko, Alan~S. Cowen, Gaurav
  Nemade, and Sujith Ravi. 2020.
\newblock \href {https://arxiv.org/abs/2005.00547} {{GoEmotions}: {A} {Dataset}
  of {Fine}-{Grained} {Emotions}}.
\newblock \emph{CoRR}, abs/2005.00547.
\newblock ArXiv: 2005.00547.

\bibitem[{Dijkstra et~al.(1995)Dijkstra, Zwaan, Graesser, and
  Magliano}]{dijkstra_character_1995}
Katinka Dijkstra, Rolf~A Zwaan, Arthur~C Graesser, and Joseph~P Magliano. 1995.
\newblock Character and reader emotions in literary texts.
\newblock \emph{Poetics}, 23(1-2):139--157.
\newblock Publisher: Elsevier.

\bibitem[{Dyer(1983)}]{dyer_role_1983}
Michael~G Dyer. 1983.
\newblock The role of affect in narratives.
\newblock \emph{Cognitive science}, 7(3):211--242.
\newblock Publisher: Wiley Online Library.

\bibitem[{Ehret et~al.(2023)Ehret, Berdicevskis, Bentz, and
  Blumenthal-Dram{\'e}}]{ehret2023measuring}
Katharina Ehret, Aleksandrs Berdicevskis, Christian Bentz, and Alice
  Blumenthal-Dram{\'e}. 2023.
\newblock Measuring language complexity: challenges and opportunities.
\newblock \emph{Linguistics Vanguard}, 9(s1):1--8.

\bibitem[{Ekman(1992)}]{ekman_argument_1992}
Paul Ekman. 1992.
\newblock \href {https://doi.org/10.1080/02699939208411068} {An argument for
  basic emotions}.
\newblock \emph{Cognition and Emotion}, 6(3-4):169--200.
\newblock Publisher: Routledge \_eprint:
  https://doi.org/10.1080/02699939208411068.

\bibitem[{Etienne(2023)}]{etienne2023analyse}
Aline Etienne. 2023.
\newblock \emph{Analyse automatique des {\'e}motions dans les textes:
  contributions th{\'e}oriques et applicatives dans le cadre de l'{\'e}tude de
  la complexit{\'e} des textes pour enfants}.
\newblock Ph.D. thesis, Universit{\'e} de Nanterre-Paris X.

\bibitem[{Etienne et~al.(2022)Etienne, Battistelli, and
  Lecorvé}]{etienne_psycho-linguistically_2022}
Aline Etienne, Delphine Battistelli, and Gw\'enol\'e Lecorvé. 2022.
\newblock \href {https://hal.science/hal-03709860} {A
  ({Psycho}-){Linguistically} {Motivated} {Scheme} for {Annotating} and
  {Exploring} {Emotions} in a {Genre}-{Diverse} {Corpus}}.
\newblock In \emph{Proceedings of the {Conference} on {Language} {Resources}
  and {Evaluation} ({LREC})}, Marseille, France.

\bibitem[{Fraisse and Paroubek(2015)}]{fraisse_utiliser_2015}
Amel Fraisse and Patrick Paroubek. 2015.
\newblock Utiliser les interjections pour détecter les émotions.
\newblock In \emph{Actes de la conférence sur le {Traitement} {Automatique}
  des {Langues} {Naturelles}}, pages 279--290.

\bibitem[{Kim and Klinger(2018)}]{kim_who_2018}
Evgeny Kim and Roman Klinger. 2018.
\newblock Who feels what and why? annotation of a literature corpus with
  semantic roles of emotions.
\newblock In \emph{Proceedings of the {International} {Conference} on
  {Computational} {Linguistics} (COLING)}, pages 1345--1359.

\bibitem[{Kim and Klinger(2019)}]{kim_analysis_2019}
Evgeny Kim and Roman Klinger. 2019.
\newblock \href {https://doi.org/10.18653/v1/W19-3406} {An {Analysis} of
  {Emotion} {Communication} {Channels} in {Fan}-{Fiction}: {Towards}
  {Emotional} {Storytelling}}.
\newblock In \emph{Proceedings of the {Workshop} on {Storytelling}}, pages
  56--64, Florence, Italy. Association for Computational Linguistics.

\bibitem[{Klinger(2023)}]{klinger2023eventcentric}
Roman Klinger. 2023.
\newblock \href {http://arxiv.org/abs/2309.02092} {Where are we in
  event-centric emotion analysis? bridging emotion role labeling and
  appraisal-based approaches}.

\bibitem[{Liu et~al.(2019)Liu, Osama, and Andrade}]{liu_dens_2019}
Chen Liu, Muhammad Osama, and Anderson~de Andrade. 2019.
\newblock \href {http://arxiv.org/abs/1910.11769} {{DENS}: {A} {Dataset} for
  {Multi}-class {Emotion} {Analysis}}.
\newblock \emph{CoRR}, abs/1910.11769.
\newblock ArXiv: 1910.11769.

\bibitem[{Martin et~al.(2020)Martin, Muller, Ortiz~Suárez, Dupont, Romary,
  de~la Clergerie, Seddah, and Sagot}]{martin_camembert_2020}
Louis Martin, Benjamin Muller, Pedro~Javier Ortiz~Suárez, Yoann Dupont,
  Laurent Romary, Éric de~la Clergerie, Djamé Seddah, and Benoît Sagot.
  2020.
\newblock \href {https://www.aclweb.org/anthology/2020.acl-main.645}
  {{CamemBERT}: a {Tasty} {French} {Language} {Model}}.
\newblock In \emph{Proceedings of the {Annual} {Meeting} of the {Association}
  for {Computational} {Linguistics} (ACL)}, pages 7203--7219, Online.
  Association for Computational Linguistics.

\bibitem[{Mohammad(2012)}]{mohammad_emotional_2012}
Saif Mohammad. 2012.
\newblock \href {https://aclanthology.org/S12-1033} {\#{Emotional} {Tweets}}.
\newblock In \emph{Proceedings of the {Joint} {Conference} on {Lexical} and
  {International} {Workshop} on {Semantic} {Evaluation} ({SemEval})}, pages
  246--255, Montréal, Canada. Association for Computational Linguistics.

\bibitem[{Mohammad et~al.(2018)Mohammad, Bravo-Marquez, Salameh, and
  Kiritchenko}]{mohammad_semeval-2018_2018}
Saif Mohammad, Felipe Bravo-Marquez, Mohammad Salameh, and Svetlana
  Kiritchenko. 2018.
\newblock \href {https://doi.org/10.18653/v1/S18-1001} {{SemEval}-2018 {Task}
  1: {Affect} in {Tweets}}.
\newblock In \emph{Proceedings of the {International} {Workshop} on {Semantic}
  {Evaluation} ({SemEval})}, pages 1--17, New Orleans, Louisiana. Association
  for Computational Linguistics.

\bibitem[{Ouyang et~al.(2022)Ouyang, Wu, Jiang, Almeida, Wainwright, Mishkin,
  Zhang, Agarwal, Slama, Ray et~al.}]{ouyang2022training}
Long Ouyang, Jeffrey Wu, Xu~Jiang, Diogo Almeida, Carroll Wainwright, Pamela
  Mishkin, Chong Zhang, Sandhini Agarwal, Katarina Slama, Alex Ray, et~al.
  2022.
\newblock Training language models to follow instructions with human feedback.
\newblock \emph{Proceedings of the Advances in Neural Information Processing
  Systems}, 35:27730--27744.

\bibitem[{Piolat and Bannour(2009)}]{piolat2009example}
Annie Piolat and Rachid Bannour. 2009.
\newblock An example of text analysis software (emotaix-tropes) use: The
  influence of anxiety on expressive writing.
\newblock \emph{Current psychology letters. Behaviour, brain \& cognition},
  25(2, 2009).

\bibitem[{Plutchik(1980)}]{plutchik_general_1980}
Robert Plutchik. 1980.
\newblock A general psychoevolutionary theory of emotion.
\newblock In \emph{Theories of emotion}, pages 3--33. Elsevier.

\bibitem[{Poria et~al.(2018)Poria, Hazarika, Majumder, Naik, Cambria, and
  Mihalcea}]{poria2018meld}
Soujanya Poria, Devamanyu Hazarika, Navonil Majumder, Gautam Naik, Erik
  Cambria, and Rada Mihalcea. 2018.
\newblock Meld: A multimodal multi-party dataset for emotion recognition in
  conversations.
\newblock \emph{arXiv preprint arXiv:1810.02508}.

\bibitem[{Poria et~al.(2019)Poria, Majumder, Mihalcea, and
  Hovy}]{poria2019emotion}
Soujanya Poria, Navonil Majumder, Rada Mihalcea, and Eduard Hovy. 2019.
\newblock Emotion recognition in conversation: Research challenges, datasets,
  and recent advances.
\newblock \emph{IEEE Access}, 7:100943--100953.

\bibitem[{Reimers and Gurevych(2019)}]{reimers2019sentence}
Nils Reimers and Iryna Gurevych. 2019.
\newblock Sentence-bert: Sentence embeddings using siamese bert-networks.
\newblock In \emph{Proceedings of the Conference on Empirical Methods in
  Natural Language Processing and the International Joint Conference on Natural
  Language Processing (EMNLP-IJCNLP)}, pages 3982--3992.

\bibitem[{Scherer(2005)}]{scherer_what_2005}
Klaus~R Scherer. 2005.
\newblock What are emotions? {And} how can they be measured?
\newblock \emph{Social science information}, 44(4):695--729.
\newblock Publisher: Sage Publications Sage CA: Thousand Oaks, CA.

\bibitem[{Strapparava and Mihalcea(2007)}]{strapparava_semeval-2007_2007}
Carlo Strapparava and Rada Mihalcea. 2007.
\newblock \href {https://aclanthology.org/S07-1013} {{SemEval}-2007 {Task} 14:
  {Affective} {Text}}.
\newblock In \emph{Proceedings of the {International} {Workshop} on {Semantic}
  {Evaluations} ({SemEval})}, pages 70--74, Prague, Czech Republic. Association
  for Computational Linguistics.

\bibitem[{Su{\'a}rez et~al.(2019)Su{\'a}rez, Sagot, and
  Romary}]{suarez2019asynchronous}
Pedro Javier~Ortiz Su{\'a}rez, Beno{\^\i}t Sagot, and Laurent Romary. 2019.
\newblock Asynchronous pipeline for processing huge corpora on medium to low
  resource infrastructures.
\newblock In \emph{Proceedings of the Workshop on the Challenges in the
  Management of Large Corpora}. Leibniz-Institut f{\"u}r Deutsche Sprache.

\bibitem[{Troiano et~al.(2023)Troiano, Oberländer, and
  Klinger}]{Troiano_klinger_2023}
E.~Troiano, L.~Oberländer, and R.~Klinger. 2023.
\newblock Dimensional modeling of emotions in text with appraisal theories:
  Corpus creation, annotation reliability, and prediction.
\newblock \emph{Computational Linguistics}, 49(1).

\bibitem[{Öhman(2020)}]{ohman_emotion_2020}
Emily Öhman. 2020.
\newblock Emotion annotation: {Rethinking} emotion categorization.
\newblock \emph{Proceedings of the CEUR Workshop}, 2865:134--144.
\newblock Publisher: CEUR-WS.

\bibitem[{Öhman et~al.(2020)Öhman, Pàmies, Kajava, and
  Tiedemann}]{ohman_xed_2020}
Emily Öhman, Marc Pàmies, Kaisla Kajava, and Jörg Tiedemann. 2020.
\newblock \href {https://doi.org/10.18653/v1/2020.coling-main.575} {{XED}: {A}
  {Multilingual} {Dataset} for {Sentiment} {Analysis} and {Emotion}
  {Detection}}.
\newblock In \emph{Proceedings of the {International} {Conference} on
  {Computational} {Linguistics} (COLING)}, pages 6542--6552, Barcelona, Spain
  (Online). International Committee on Computational Linguistics.

\end{thebibliography}

\appendix

\appendix

\section{Inter-Annotator Agreement}
\label{app:iaa}

To maximize the number of annotations, each text has been annotated by one annotator. Then, the validity of the annotations has been evaluated by comparing the annotations of two productive annotators from the 6~ones in the whole campaign (referred to as A1 and A2) with another 7-th expert. The Cohen's Kappa for each label is given in~\ref{tab:iaa}. For a comparison, in \cite{kim_who_2018}, \textit{joy} is annotated with a Kappa value of~$0.4$.

\begin{table}[t!]
\begin{center}
\includegraphics[scale=0.8]{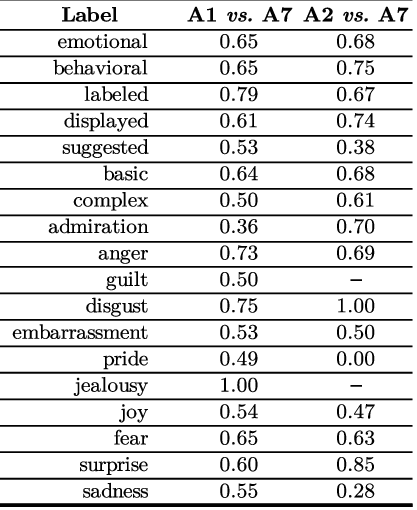} 
\caption{Cohen's Kappa for each label between annotators A1 and A2, and an additional one, A7}
\label{tab:iaa}
\end{center}
\vspace{-5mm}
\end{table}

\section{Confusion matrices}

Table~\ref{tab:confusion_matrices} reports the confusion matrices for each task for our model.

These details first show that, inspite of the imbalance between emotional and non-emotional samples, the classification is not biased.

Regarding the modes (B), most frequent errors relates to guessing a mode where there is none or the contrary. Then, \textit{behavioral} and \textit{suggested} modes are those with the highest number of false positives. This is probably due to the fact that these mode are less direct and require interpretation. Finally, the most frequent confusions between two modes are \textit{labeled} guessed as \textit{suggested}, and the reciprocal. This maybe means that the definition of a \textit{suggested} emotion falls back in the end to a valence consideration of lexical items.

On the side of types (C), the biggest confusion is with the \textit{none} case. Then, there is no specific bias towards \textit{basic} or \textit{complex}.

Finally, our model primarily predicts the classes of Task D where they are expected. Mainly, no prevalent confusion between classes emerges. When no emotional category is expected, the classes \textit{other,} \textit{anger,} \textit{fear,} and \textit{sadness} are the most predicted. These false positives do not clash with the \textit{other} class, which is very heterogeneous, but are surprising for the other classes, which are normally better defined.

\begin{table}[t!]
\begin{center}
\includegraphics[scale=0.8]{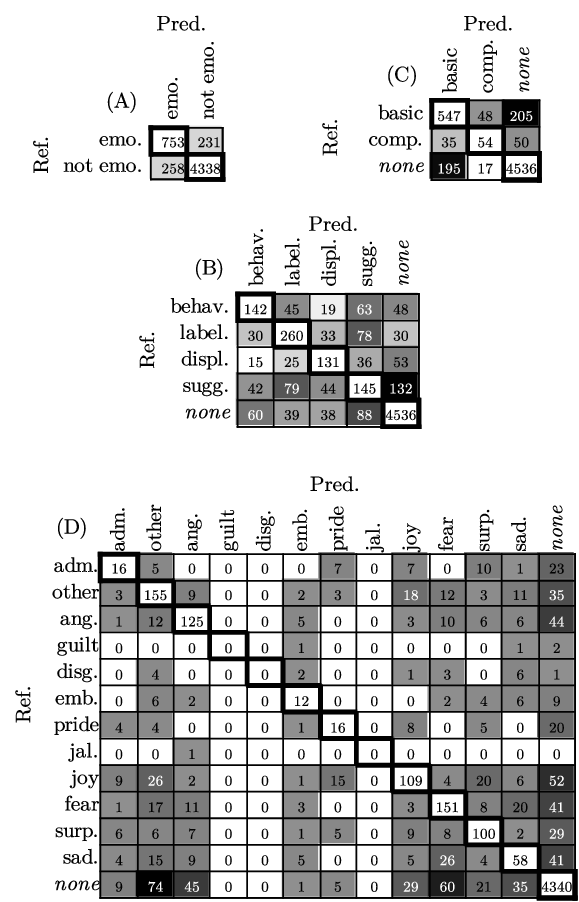} 
\caption{Confusion matrices of our model for Tasks A, B, C, and~D}
\label{tab:confusion_matrices}
\end{center}
\vspace{-5mm}
\end{table}

\section{Prompts for GPT-3.5}
\label{app:gpt35}

GPT-3.5 was used in conversational mode. The prompts are thus an alternation of messages between the \textit{user} and the \textit{assistant}, preceded by a global message from the \textit{system}. The user's messages cover all the labels from all tasks A to D, explaining the meaning of each label, while the assistant's responses are binary ("yes" / "no") to indicate the presence or absence of the respective class. Two types of messages are considered for the user: either the explanations of each label are accompanied by positive examples, or they are accompanied by both positive and negative examples (i.e., counter-examples). Only Task A (presence or absence of emotional information) is an exception since it is always accompanied by counter-examples, regardless of the type of prompt. Sections~\ref{subsec:details_sans_contre_exemples} and~\ref{subsec:details_avec_contre_exemples} then show the details of the two types of prompts. We use version 0311 of GPT-3.5 for all experiments.

\subsection{Detailed results}
\label{subsec:details_resultats_gpt35}

Table~\ref{tab:details_gpt35} provides detailed results on the test set for each approach compared to our model. Overall, these results show that our model performs better and that the approach without counter-examples is better than the one with counter-examples. The main problem with GPT-3.5 seems to be that it predicts too many labels (high recall but low precision). However, it is worth noting that GPT-3.5 seems to perform better on rare classes because our model does not predict them.

\begin{table*}[t!]
\begin{center}
\includegraphics[scale=0.74]{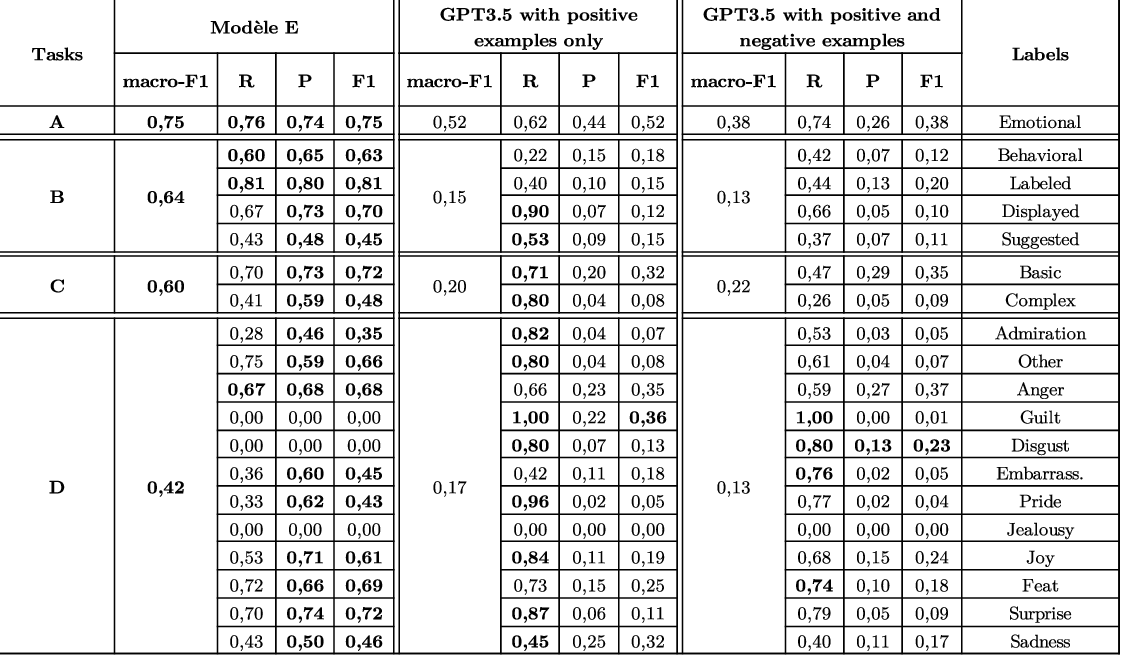} 
\caption{Detailed comparison between our model and the two approaches based on GPT3.5}
\label{tab:details_gpt35}
\end{center}
\vspace{-5mm}
\end{table*}

\subsection{With positive examples only}
\label{subsec:details_sans_contre_exemples}

{\tt\scriptsize
\textbf{System:}\\
Tu joues le rôle d'un expert linguiste qui annote des phrases en t'intéressant à leur dimension émotionnelle.

L'annotation porte au niveau de la phrase et prend la forme de questions successives. Pour comprendre le contexte, la phrase à annoter est donnée avec sa phrase précédente et sa phrase suivante, mais la réponse à chaque question doit uniquement porter sur la seule phrase à annoter, et non sur la phrase précédente ou suivante.

- Phrase précédente: Nicolas Hulot n’appartient à aucun parti politique.\\
- Phrase à annoter: Il a refusé trois fois le poste de ministre de l’Ecologie avant d’accepter la proposition d’Emmanuel Macron.\\
- Phrase suivante: Mais ça ne s’est pas très bien passé.

\textbf{User:}\\
Définition: une phrase est dite "émotionnelle" si elle exprime explicitement ou implicitement une émotion, qu'elle soit exprimée par le narrateur ou un personnage. Par exemple:\\
- émotionnelle: "Cette information a beaucoup énervé Marie."\\
- émotionnelle: "Andrée a sautillé partout en chantant."\\
- émotionnelle: "Oh, non... C'est vraiment dommage !"\\
- émotionnelle: "Ces deux amis se retrouvent après une longue séparation."\\
- non émotionnelle: "Avant d'arriver devant une salle de classe, les enseignants, eux aussi, sont sur les bancs de l'école."\\
- non émotionnelle: "De 2007 à 2012, il a été le Premier ministre de l'ancien président Nicolas Sarkozy."\\
- non émotionnelle: "Récemment, une nouvelle autorisation a été délivrée pour un deuxième test dans le courant de l'année 2019."\\
- non émotionnelle: "Avant de sortir, Billy prépare un dîner orange : une soupe de potiron, des cuisses de canard à l'orange avec une purée de carottes et une tarte à la citrouille."

Question: La phrase à annoter est-elle **émotionnelle** ?

Réponse (oui/non):

\textbf{Assistant:}\\
\textit{<réponse du modèle>}

\textbf{User:}\\
Définition: La catégorie émotionnelle "colère" recouvre les émotions suivantes: agacement, colère, contestation, désaccord (si émotion suggérée), désapprobation, énervement, fureur/rage, indignation, insatisfaction, irritation, mécontentement, réprobation et révolte. Par exemple :\\
- "C'est notamment pour cette raison que des "gilets jaunes", les personnes qui manifestent et bloquent des routes dans le pays depuis plusieurs semaines, sont en colère."\\
- "- Ton commentaire est déplacé, jeune homme ! a-t-elle dit d'un air pincé."

Question: Si la phrase à annoter est émotionnelle, est-ce que la catégorie émotionnelle **colère** est présente ?

Réponse (oui/non):

\textbf{Assistant:}\\
\textit{<réponse du modèle>}

\textbf{User:}\\
Définition: La catégorie émotionnelle "dégoût" recouvre les émotions suivantes: dégoût, lassitude et répulsion. Par exemple :\\
- "Beurk !"\\
- "Ça peut paraître dégoûtant, mais on peut manger des insectes."\\

Question: Si la phrase à annoter est émotionnelle, est-ce que la catégorie émotionnelle **dégoût** est présente ?

Réponse (oui/non):

\textbf{Assistant:}\\
\textit{<réponse du modèle>}

\textbf{User:}\\
Définition: La catégorie émotionnelle "joie" recouvre les émotions suivantes: amusement, enthousiasme, exaltation, joie et plaisir. Par exemple :\\
- "Pour fêter ses buts, il lui arrive souvent de danser."\\
- "- Je suis bien aise de vous voir, me dit le roi sur un ton amical."

Question: Si la phrase à annoter est émotionnelle, est-ce que la catégorie émotionnelle **joie** est présente ?

Réponse (oui/non):

\textbf{Assistant:}\\
\textit{<réponse du modèle>}

\textbf{User:}\\
Définition: La catégorie émotionnelle "peur" recouvre les émotions suivantes: angoisse, appréhension, effroi, horreur, inquiétude, méfiance, peur, stress et timidité. Par exemple :\\
- "Le Front national, qui est d'extrême droite, faisait peur, à cause des idées qu'il défendait."\\
- "Il y avait un grand silence dans la maison."

Question: Si la phrase à annoter est émotionnelle, est-ce que la catégorie émotionnelle **peur** est présente ?

Réponse (oui/non):

\textbf{Assistant:}\\
\textit{<réponse du modèle>}

\textbf{User:}\\
Définition: La catégorie émotionnelle "surprise" recouvre les émotions suivantes: étonnement, stupeur, surprise. Par exemple :\\
- "Finalement, ils ont été pris en charge... par les agriculteurs locaux, dans un camion benne !"\\
- "Tous, étonnés, se taisent."

Question: Si la phrase à annoter est émotionnelle, est-ce que la catégorie émotionnelle **surprise** est présente ?

Réponse (oui/non):

\textbf{Assistant:}\\
\textit{<réponse du modèle>}

\textbf{User:}\\
Définition: La catégorie émotionnelle "tristesse" recouvre les émotions suivantes: blues, chagrin, déception, désespoir, peine, souffrance et tristesse. Par exemple :\\
- "Sa mère venait de mourir et son père était au front."\\
- "L'âne continuait à examiner la peinture d'un regard plutôt attristé."

Question: Si la phrase à annoter est émotionnelle, est-ce que la catégorie émotionnelle **tristesse** est présente ?

Réponse (oui/non):

\textbf{Assistant:}\\
\textit{<réponse du modèle>}

\textbf{User:}\\
Définition: La catégorie émotionnelle "admiration" recouvre les émotions suivantes: admiration. Par exemple :\\
- "De nos jours, ce site exceptionnel permet de montrer toute la richesse de la civilisation romaine et la façon dont les villes et la société étaient organisées."\\
- "- Tes enfants sont vraiment merveilleux, ma chérie, dit-elle à sa fille."

Question: Si la phrase à annoter est émotionnelle, est-ce que la catégorie émotionnelle **admiration** est présente ?

Réponse (oui/non):

\textbf{Assistant:}\\
\textit{<réponse du modèle>}

\textbf{User:}\\
Définition: La catégorie émotionnelle "culpabilité" recouvre les émotions suivantes: culpabilité. Par exemple :\\
- "Et je l'avais bien mérité."\\
- "Surtout, il ne faut pas se sentir coupable de ne pas avoir réagi."

Question: Si la phrase à annoter est émotionnelle, est-ce que la catégorie émotionnelle **culpabilité** est présente ?

Réponse (oui/non):

\textbf{Assistant:}\\
\textit{<réponse du modèle>}

\textbf{User:}\\
Définition: La catégorie émotionnelle "embarras" recouvre les émotions suivantes: embarras, gêne, honte, humiliation et timidité. Par exemple :\\
- "Après cette humiliante défaite, Napoléon abdique une nouvelle fois, ce qui marque définitivement la fin de l'Empire et de sa période de retour appelée ''les Cent jours''."\\
- "Légèrement décontenancée, la prof s'est raclé la gorge et commencé la lecture."

Question: Si la phrase à annoter est émotionnelle, est-ce que la catégorie émotionnelle **embarras** est présente ?

Réponse (oui/non):

\textbf{Assistant:}\\
\textit{<réponse du modèle>}

\textbf{User:}\\
Définition: La catégorie émotionnelle "fierté" recouvre les émotions suivantes: fierté et orgueil. Par exemple :\\
- "Flavia entre dans la cour comme une conquérante, entourée de ses supporters."\\
- "Magawa peut être fier de lui, car il vient de recevoir une médaille d'or."

Question: Si la phrase à annoter est émotionnelle, est-ce que la catégorie émotionnelle **fierté** est présente ?

Réponse (oui/non):

\textbf{Assistant:}\\
\textit{<réponse du modèle>}

\textbf{User:}\\
Définition: La catégorie émotionnelle "jalousie" recouvre les émotions suivantes: jalousie. Par exemple :\\
- "Mais quand Flavia découvre le jeune génie du piano, elle se sent comme écrasée."\\
- "On dirait presque qu'il fait partie de l'instrument."

Question: Si la phrase à annoter est émotionnelle, est-ce que la catégorie émotionnelle **jalousie** est présente ?

Réponse (oui/non):

\textbf{Assistant:}\\
\textit{<réponse du modèle>}

\textbf{User:}\\
Définition: La catégorie émotionnelle "autre" recouvre les émotions suivantes: amour, courage, curiosité, désir, détermination, envie, espoir, haine, impuissance, mépris et soulagement. Par exemple :\\
- "Dans chaque camp, ils se sont mobilisés pour donner envie aux gens de voter comme eux."\\
- "Ils n'apprécient pas du tout l'attitude des dirigeants, notamment celle du président, ''qu'ils jugent méprisant, déconnecté de la réalité, du quotidien'', note le sociologue Alexis Spire."

Question: Si la phrase à annoter est émotionnelle, est-ce que la catégorie émotionnelle **autre** est présente ?

Réponse (oui/non):

\textbf{Assistant:}\\
\textit{<réponse du modèle>}

\textbf{User:}\\
Définition: Les émotions suivantes sont dites "de base" : Colère, Dégoût, Joie, Peur, Surprise, Tristesse.

Question: Si la phrase à annoter est émotionnelle, contient-elle une **émotion de base** ?

Réponse (oui/non):

\textbf{Assistant:}\\
\textit{<réponse du modèle>}

\textbf{User:}\\
Définition: Les émotions suivantes sont dites "complexes": Admiration, Culpabilité, Embarras, Fierté, Jalousie.

Question: Si la phrase à annoter est émotionnelle, contient-elle une **émotion complexe** ?

Réponse (oui/non):

\textbf{Assistant:}\\
\textit{<réponse du modèle>}

\textbf{User:}\\
Définition: Une émotion est dite du mode "désigné" lorsqu'elle est exprimée par un terme du lexique émotionnel. Par exemple :\\
- "Pierre est heureux d'être bientôt à la retraite.", où la joie de Pierre est désignée par le terme "heureux".\\
- "Cette information a beaucoup énervé Marie.", où la colère de Marie est désignée par le terme "énervé".

Question: Si la phrase à annoter est émotionnelle, est-ce que le mode **désigné** est utilisé ?

Réponse (oui/non):

\textbf{Assistant:}\\
\textit{<réponse du modèle>}

\textbf{User:}\\
Définition: Une émotion est dite du mode "comportemental" lorsqu'elle est exprimée par la description d'une manifestation physique (physiologique ou comportementale) de l'émotion. Par exemple :\\
- "Paul sanglote.", où la tristesse de Paul est exprimée par le comportement "sanglote".\\
- "Andrée a sautillé partout en chantant.", où la joie de Andrée est exprimée par le comportement "sautillé partout en chantant".

Question: Si la phrase à annoter est émotionnelle, est-ce que le mode **comportemental** est utilisé ?

Réponse (oui/non):

\textbf{Assistant:}\\
\textit{<réponse du modèle>}

\textbf{User:}\\
Définition: Une émotion est dite du mode "montré" lorsqu'elle est exprimée par des caractéristiques linguistiques de l'énoncé qui traduisent l'état émotionnel dans lequel se trouvait l'énonciateur au moment de l'énonciation. Par exemple :\\
- "Oh, chouette ! Quelle bonne idée !", car la joie de l'énonciateur est traduite au sein de l'énoncé par les interjections "oh" et "chouette", les énoncés averbaux et les points d'exclamations.\\
- "Oh, non... C'est vraiment dommage !", car la tristesse de l'énonciateur est traduite au sein de l'énoncé par l'interjection "oh", l'énoncé averbal, les points de suspension et le point d'exclamation.

Question: Si la phrase à annoter est émotionnelle, est-ce que le mode **montré** est utilisé ?

Réponse (oui/non):

\textbf{Assistant:}\\
\textit{<réponse du modèle>}

\textbf{User:}\\
Définition: Une émotion est dite du mode "suggéré" lorsqu'elle est exprimée par la description d'une situation associée de manière conventionnelle à un ressenti émotionnel. Par exemple :\\
- "Le père de Jeanne est mort hier à cause d'un cancer.", où la tristesse de Jeanne est suggérée par la description du décès, il y a peu de temps, de son père (une personne proche d'elle).\\
- "Ces deux amis se retrouvent après une longue séparation.", où la joie des deux amis est suggérée par la description de leurs retrouvailles après un temps long.

Question: Si la phrase à annoter est émotionnelle, est-ce que le mode **suggéré** est utilisé ?

Réponse (oui/non):

\textbf{Assistant:}\\
\textit{<réponse du modèle>}
}

\subsection{With positive and negative examples}
\label{subsec:details_avec_contre_exemples}

{\tt\scriptsize
\textbf{System:}\\
Tu joues le rôle d'un expert linguiste qui annote des phrases d'après leurs dimensions émotionnelle.\\

Les différentes annotations sont toute binaires (absence ou présence d'une propriété). Elles vont porter sur la nature émotionnelle ou non des phrases et, si oui, le mode d'expression de la ou des émotions présentes (désignée, comportementale, montrée ou suggérée), la ou les catégories émotionnelles (joie, peur, colère, tristesse, etc.) et le ou les types d'émotion ("de base" ou "complexe"). Chaque propriété est décrite par une définition et des exemples.\\

L'annotation La phrase à annoter est entourée des balises <annotate>...</annotate>.\\

\textbf{User:}\\
Définition : une phrase est dite "émotionnelle" si elle exprime explicitement ou implicitement une émotion, qu'elle soit exprimée par le narrateur ou un personnage.\\

Question : La phrase à annoter est-elle **émotionnelle** ?\\

Exemples :\\\\
- <annotate>Avant de sortir, Billy prépare un dîner orange : une soupe de potiron, des cuisses de canard à l'orange avec une purée de carottes et une tarte à la citrouille.</annotate> -> non\\
- <annotate>Cette information a beaucoup énervé Marie.</annotate> -> oui\\
- <annotate>Andrée a sautillé partout en chantant.</annotate> -> oui\\
- <annotate>Récemment, une nouvelle autorisation a été délivrée pour un deuxième test dans le courant de l'année 2019.</annotate> -> non
- <annotate>Oh, non... C'est vraiment dommage !</annotate> -> oui\\
- <annotate>De 2007 à 2012, il a été le Premier ministre de l'ancien président Nicolas Sarkozy.</annotate> -> non\\
- <annotate>Ces deux amis se retrouvent après une longue séparation. -> oui\\
- <annotate>Avant d'arriver devant une salle de classe, les enseignants, eux aussi, sont sur les bancs de l'école.</annotate> -> non\\

Annotation (oui/non) :\\
- Nicolas Hulot n’appartient à aucun parti politique. <annotate>Il a refusé trois fois le poste de ministre de l’Ecologie avant d’accepter la proposition d’Emmanuel Macron.</annotate> Mais ça ne s’est pas très bien passé.  -> \\

\textbf{Assistant:}\\
\textit{réponse du modèle}\\

\textbf{User:}\\
Définition : La catégorie émotionnelle "colère" recouvre les émotions suivantes: agacement, colère, contestation, désaccord (si émotion suggérée), désapprobation, énervement, fureur/rage, indignation, insatisfaction, irritation, mécontentement, réprobation et révolte.\\

Question : Si la phrase à annoter est émotionnelle, est-ce que la catégorie émotionnelle **colère** est présente ?\\

Exemples :\\
- <annotate>De 2007 à 2012, il a été le Premier ministre de l'ancien président Nicolas Sarkozy.</annotate> -> non\\
- <annotate>C'est notamment pour cette raison que des "gilets jaunes", les personnes qui manifestent et bloquent des routes dans le pays depuis plusieurs semaines, sont en colère.</annotate> -> oui.\\
- <annotate>Tous, étonnés, se taisent.</annotate> -> non.\\
- <annotate>- Ton commentaire est déplacé, jeune homme ! a-t-elle dit d'un air pincé.</annotate> -> oui.\\
- <annotate>Après cette humiliante défaite, Napoléon abdique une nouvelle fois, ce qui marque définitivement la fin de l'Empire et de sa période de retour appelée "les Cent jours".</annotate> -> non.\\

Annotation (oui/non) :\\
- Nicolas Hulot n’appartient à aucun parti politique. <annotate>Il a refusé trois fois le poste de ministre de l’Ecologie avant d’accepter la proposition d’Emmanuel Macron.</annotate> Mais ça ne s’est pas très bien passé.  -> \\

\textbf{Assistant:}\\
\textit{réponse du modèle}\\

\textbf{User:}\\
Définition : La catégorie émotionnelle "dégoût" recouvre les émotions suivantes: dégoût, lassitude et répulsion.\\

Question : Si la phrase à annoter est émotionnelle, est-ce que la catégorie émotionnelle **dégoût** est présente ?\\

Exemples :\\
- <annotate>Ça peut paraître dégoûtant, mais on peut manger des insectes.</annotate> -> oui.\\
- <annotate>Beurk !</annotate> -> oui.\\
- <annotate>Finalement, ils ont été pris en charge... par les agriculteurs locaux, dans un camion benne !</annotate> -> non.\\
- <annotate>Le Front national, qui est d'extrême droite, faisait peur, à cause des idées qu'il défendait.</annotate> -> non.\\
- <annotate>Avant d'arriver devant une salle de classe, les enseignants, eux aussi, sont sur les bancs de l'école.</annotate> -> non\\

Annotation (oui/non) :\\
- Nicolas Hulot n’appartient à aucun parti politique. <annotate>Il a refusé trois fois le poste de ministre de l’Ecologie avant d’accepter la proposition d’Emmanuel Macron.</annotate> Mais ça ne s’est pas très bien passé.  -> \\

\textbf{Assistant:}\\
\textit{réponse du modèle}\\

\textbf{User:}\\
Définition : La catégorie émotionnelle "joie" recouvre les émotions suivantes: amusement, enthousiasme, exaltation, joie et plaisir.\\

Question : Si la phrase à annoter est émotionnelle, est-ce que la catégorie émotionnelle **joie** est présente ?\\

Exemples :\\
- <annotate>Dans chaque camp, ils se sont mobilisés pour donner envie aux gens de voter comme eux.</annotate> -> non.\\
- <annotate>- Je suis bien aise de vous voir, me dit le roi sur un ton amical.</annotate> -> oui.\\
- <annotate>Beurk !</annotate> -> non.\\
- <annotate>Avant d'arriver devant une salle de classe, les enseignants, eux aussi, sont sur les bancs de l'école.</annotate> -> non\\
- <annotate>Pour fêter ses buts, il lui arrive souvent de danser.</annotate> -> oui.\\

Annotation (oui/non) :\\
- Nicolas Hulot n’appartient à aucun parti politique. <annotate>Il a refusé trois fois le poste de ministre de l’Ecologie avant d’accepter la proposition d’Emmanuel Macron.</annotate> Mais ça ne s’est pas très bien passé.  -> \\

\textbf{Assistant:}\\
\textit{réponse du modèle}\\

\textbf{User:}\\
Définition : La catégorie émotionnelle "peur" recouvre les émotions suivantes: angoisse, appréhension, effroi, horreur, inquiétude, méfiance, peur, stress et timidité.\\

Question : Si la phrase à annoter est émotionnelle, est-ce que la catégorie émotionnelle **peur** est présente ?\\

Exemples :\\
- <annotate>Le Front national, qui est d'extrême droite, faisait peur, à cause des idées qu'il défendait.</annotate> -> oui.\\
- <annotate>Dans chaque camp, ils se sont mobilisés pour donner envie aux gens de voter comme eux.</annotate> -> non.\\
- <annotate>Ça peut paraître dégoûtant, mais on peut manger des insectes.</annotate> -> non.\\
- <annotate>Récemment, une nouvelle autorisation a été délivrée pour un deuxième test dans le courant de l'année 2019.</annotate> -> non\\
- <annotate>Il y avait un grand silence dans la maison.</annotate> -> oui.\\

Annotation (oui/non) :\\
- Nicolas Hulot n’appartient à aucun parti politique. <annotate>Il a refusé trois fois le poste de ministre de l’Ecologie avant d’accepter la proposition d’Emmanuel Macron.</annotate> Mais ça ne s’est pas très bien passé.  -> \\

\textbf{Assistant:}\\
\textit{réponse du modèle}\\

\textbf{User:}\\
Définition : La catégorie émotionnelle "surprise" recouvre les émotions suivantes: étonnement, stupeur, surprise.\\

Question : Si la phrase à annoter est émotionnelle, est-ce que la catégorie émotionnelle **surprise** est présente ?\\

Exemples :\\
- <annotate>Finalement, ils ont été pris en charge... par les agriculteurs locaux, dans un camion benne !</annotate> -> oui.\\
- <annotate>Avant d'arriver devant une salle de classe, les enseignants, eux aussi, sont sur les bancs de l'école.</annotate> -> non\\
- <annotate>Mais quand Flavia découvre le jeune génie du piano, elle se sent comme écrasée.</annotate> -> non.\\
- <annotate>Beurk !</annotate> -> non.\\
- <annotate>Tous, étonnés, se taisent.</annotate> -> oui.\\

Annotation (oui/non) :\\
- Nicolas Hulot n’appartient à aucun parti politique. <annotate>Il a refusé trois fois le poste de ministre de l’Ecologie avant d’accepter la proposition d’Emmanuel Macron.</annotate> Mais ça ne s’est pas très bien passé.  -> \\

\textbf{Assistant:}\\
\textit{réponse du modèle}\\

\textbf{User:}\\
Définition : La catégorie émotionnelle "tristesse" recouvre les émotions suivantes: blues, chagrin, déception, désespoir, peine, souffrance et tristesse.\\

Question : Si la phrase à annoter est émotionnelle, est-ce que la catégorie émotionnelle **tristesse** est présente ?\\

Exemples :\\
- <annotate>Avant d'arriver devant une salle de classe, les enseignants, eux aussi, sont sur les bancs de l'école.</annotate> -> non\\
- <annotate>Le Front national, qui est d'extrême droite, faisait peur, à cause des idées qu'il défendait.</annotate> -> non.\\
- <annotate>Sa mère venait de mourir et son père était au front.</annotate> -> oui.\\
- <annotate>Légèrement décontenancée, la prof s'est raclé la gorge et commencé la lecture.</annotate> -> non.\\
- <annotate>L'âne continuait à examiner la peinture d'un regard plutôt attristé.</annotate> -> oui.

Annotation (oui/non) :\\
- Nicolas Hulot n’appartient à aucun parti politique. <annotate>Il a refusé trois fois le poste de ministre de l’Ecologie avant d’accepter la proposition d’Emmanuel Macron.</annotate> Mais ça ne s’est pas très bien passé.  -> \\

\textbf{Assistant:}\\
\textit{réponse du modèle}\\

\textbf{User:}\\
Définition : La catégorie émotionnelle "admiration" recouvre les émotions suivantes: admiration.\\

Question : Si la phrase à annoter est émotionnelle, est-ce que la catégorie émotionnelle **admiration** est présente ?\\

Exemples :\\
- <annotate>Tous, étonnés, se taisent.</annotate> -> non.\\
- <annotate>De nos jours, ce site exceptionnel permet de montrer toute la richesse de la civilisation romaine et la façon dont les villes et la société étaient organisées.</annotate> -> oui.\\
- <annotate>Magawa peut être fier de lui, car il vient de recevoir une médaille d'or.</annotate> -> non.\\
- <annotate>Avant de sortir, Billy prépare un dîner orange : une soupe de potiron, des cuisses de canard à l'orange avec une purée de carottes et une tarte à la citrouille.</annotate> -> non\\
- <annotate>- Tes enfants sont vraiment merveilleux, ma chérie, dit-elle à sa fille.</annotate> -> oui.\\

Annotation (oui/non) :\\
- Nicolas Hulot n’appartient à aucun parti politique. <annotate>Il a refusé trois fois le poste de ministre de l’Ecologie avant d’accepter la proposition d’Emmanuel Macron.</annotate> Mais ça ne s’est pas très bien passé.  -> \\

\textbf{Assistant:}\\
\textit{réponse du modèle}\\

\textbf{User:}\\
Définition : La catégorie émotionnelle "culpabilité" recouvre les émotions suivantes: culpabilité.\\

Question : Si la phrase à annoter est émotionnelle, est-ce que la catégorie émotionnelle **culpabilité** est présente ?\\

Exemples :\\
- <annotate>Et je l'avais bien mérité.</annotate> -> oui.\\
- <annotate>Tous, étonnés, se taisent.</annotate> -> non.\\
- <annotate>Surtout, il ne faut pas se sentir coupable de ne pas avoir réagi.</annotate> -> oui.\\
- <annotate>Tous, étonnés, se taisent.</annotate> -> non.\\
- <annotate>Avant d'arriver devant une salle de classe, les enseignants, eux aussi, sont sur les bancs de l'école.</annotate> -> non\\

Annotation (oui/non) :\\
- Nicolas Hulot n’appartient à aucun parti politique. <annotate>Il a refusé trois fois le poste de ministre de l’Ecologie avant d’accepter la proposition d’Emmanuel Macron.</annotate> Mais ça ne s’est pas très bien passé.  -> \\

\textbf{Assistant:}\\
\textit{réponse du modèle}\\

\textbf{User:}\\
Définition : La catégorie émotionnelle "embarras" recouvre les émotions suivantes: embarras, gêne, honte, humiliation et timidité.\\

Question : Si la phrase à annoter est émotionnelle, est-ce que la catégorie émotionnelle **embarras** est présente ?\\

Exemples :\\
- <annotate>Le Front national, qui est d'extrême droite, faisait peur, à cause des idées qu'il défendait.</annotate> -> non.\\
- <annotate>- Tes enfants sont vraiment merveilleux, ma chérie, dit-elle à sa fille.</annotate> -> non.\\
- <annotate>Avant d'arriver devant une salle de classe, les enseignants, eux aussi, sont sur les bancs de l'école.</annotate> -> non\\
- <annotate>Après cette humiliante défaite, Napoléon abdique une nouvelle fois, ce qui marque définitivement la fin de l'Empire et de sa période de retour appelée "les Cent jours".</annotate> -> oui.\\
- <annotate>Légèrement décontenancée, la prof s'est raclé la gorge et commencé la lecture.</annotate> -> oui.\\

Annotation (oui/non) :\\
- Nicolas Hulot n’appartient à aucun parti politique. <annotate>Il a refusé trois fois le poste de ministre de l’Ecologie avant d’accepter la proposition d’Emmanuel Macron.</annotate> Mais ça ne s’est pas très bien passé.  -> \\

\textbf{Assistant:}\\
\textit{réponse du modèle}\\

\textbf{User:}\\
Définition : La catégorie émotionnelle "fierté" recouvre les émotions suivantes: fierté et orgueil.\\

Question : Si la phrase à annoter est émotionnelle, est-ce que la catégorie émotionnelle **fierté** est présente ?\\

Exemples :\\
- <annotate>Avant de sortir, Billy prépare un dîner orange : une soupe de potiron, des cuisses de canard à l'orange avec une purée de carottes et une tarte à la citrouille.</annotate> -> non\\
- <annotate>On dirait presque qu'il fait partie de l'instrument.</annotate> -> non.\\
- <annotate>Magawa peut être fier de lui, car il vient de recevoir une médaille d'or.</annotate> -> oui.\\
- <annotate>Flavia entre dans la cour comme une conquérante, entourée de ses supporters.</annotate> -> oui.\\
- <annotate>Il y avait un grand silence dans la maison.</annotate> -> non.\\

Annotation (oui/non) :\\
- Nicolas Hulot n’appartient à aucun parti politique. <annotate>Il a refusé trois fois le poste de ministre de l’Ecologie avant d’accepter la proposition d’Emmanuel Macron.</annotate> Mais ça ne s’est pas très bien passé.  -> \\

\textbf{Assistant:}\\
\textit{réponse du modèle}\\

\textbf{User:}\\
Définition : La catégorie émotionnelle "jalousie" recouvre les émotions suivantes: jalousie.\\

Question : Si la phrase à annoter est émotionnelle, est-ce que la catégorie émotionnelle **jalousie** est présente ?\\

Exemples :\\
- <annotate>On dirait presque qu'il fait partie de l'instrument.</annotate> -> oui.\\
- <annotate>Et je l'avais bien mérité.</annotate> -> non.\\
- <annotate>Et je l'avais bien mérité.</annotate> -> non.\\
- <annotate>Mais quand Flavia découvre le jeune génie du piano, elle se sent comme écrasée.</annotate> -> oui.\\
- <annotate>Avant d'arriver devant une salle de classe, les enseignants, eux aussi, sont sur les bancs de l'école.</annotate> -> non\\

Annotation (oui/non) :\\
- Nicolas Hulot n’appartient à aucun parti politique. <annotate>Il a refusé trois fois le poste de ministre de l’Ecologie avant d’accepter la proposition d’Emmanuel Macron.</annotate> Mais ça ne s’est pas très bien passé.  -> \\

\textbf{Assistant:}\\
\textit{réponse du modèle}\\

\textbf{User:}\\
Définition : La catégorie émotionnelle "autre" recouvre les émotions suivantes: amour, courage, curiosité, désir, détermination, envie, espoir, haine, impuissance, mépris et soulagement.\\

Question : Si la phrase à annoter est émotionnelle, est-ce que la catégorie émotionnelle **autre** est présente ?\\

Exemples :\\
- <annotate>De nos jours, ce site exceptionnel permet de montrer toute la richesse de la civilisation romaine et la façon dont les villes et la société étaient organisées.</annotate> -> non.\\
- <annotate>L'âne continuait à examiner la peinture d'un regard plutôt attristé.</annotate> -> non.\\
- <annotate>Récemment, une nouvelle autorisation a été délivrée pour un deuxième test dans le courant de l'année 2019.</annotate> -> non\\
- <annotate>Ils n'apprécient pas du tout l'attitude des dirigeants, notamment celle du président, "qu'ils jugent méprisant, déconnecté de la réalité, du quotidien", note le sociologue Alexis Spire.</annotate> -> oui.\\
- <annotate>Dans chaque camp, ils se sont mobilisés pour donner envie aux gens de voter comme eux.</annotate> -> oui.\\

Annotation (oui/non) :\\
- Nicolas Hulot n’appartient à aucun parti politique. <annotate>Il a refusé trois fois le poste de ministre de l’Ecologie avant d’accepter la proposition d’Emmanuel Macron.</annotate> Mais ça ne s’est pas très bien passé.  -> \\

\textbf{Assistant:}\\
\textit{réponse du modèle}\\

\textbf{User:}\\
Définition : Les émotions suivantes sont dites "de base" : Colère, Dégoût, Joie, Peur, Surprise, Tristesse.\\

Question : Si la phrase à annoter est émotionnelle, contient-elle une **émotion de base** ?\\

Annotation (oui/non) :\\
- Nicolas Hulot n’appartient à aucun parti politique. <annotate>Il a refusé trois fois le poste de ministre de l’Ecologie avant d’accepter la proposition d’Emmanuel Macron.</annotate> Mais ça ne s’est pas très bien passé.  -> \\

\textbf{Assistant:}\\
\textit{réponse du modèle}\\

\textbf{User:}\\
Définition : Les émotions suivantes sont dites "complexes": Admiration, Culpabilité, Embarras, Fierté, Jalousie.\\

Question : Si la phrase à annoter est émotionnelle, contient-elle une **émotion complexe** ?\\

Annotation (oui/non) :\\
- Nicolas Hulot n’appartient à aucun parti politique. <annotate>Il a refusé trois fois le poste de ministre de l’Ecologie avant d’accepter la proposition d’Emmanuel Macron.</annotate> Mais ça ne s’est pas très bien passé.  -> \\

\textbf{Assistant:}\\
\textit{réponse du modèle}\\

\textbf{User:}\\
Définition : Une émotion est dite du mode "désigné" lorsqu'elle est exprimée par un terme du lexique émotionnel.\\

Question : Si la phrase à annoter est émotionnelle, est-ce que le mode **désigné** est utilisé ?\\

Exemples :\\
- <annotate>Pierre est heureux d'être bientôt à la retraite.</annotate> -> oui (car la joie de Pierre est désignée par le terme "heureux").\\
- <annotate>Oh, non... C'est vraiment dommage !</annotate> -> non.\\
- <annotate>Avant d'arriver devant une salle de classe, les enseignants, eux aussi, sont sur les bancs de l'école.</annotate> -> non\\
- <annotate>Oh, non... C'est vraiment dommage !</annotate> -> non.\\
- <annotate>Cette information a beaucoup énervé Marie.</annotate> -> oui (car la colère de Marie est désignée par le terme "énervé").\\

Annotation (oui/non) :\\
- Nicolas Hulot n’appartient à aucun parti politique. <annotate>Il a refusé trois fois le poste de ministre de l’Ecologie avant d’accepter la proposition d’Emmanuel Macron.</annotate> Mais ça ne s’est pas très bien passé.  -> \\

\textbf{Assistant:}\\
\textit{réponse du modèle}\\

\textbf{User:}\\
Définition : Une émotion est dite du mode "comportemental" lorsqu'elle est exprimée par la description d'une manifestation physique (physiologique ou comportementale) de l'émotion.\\

Question : Si la phrase à annoter est émotionnelle, est-ce que le mode **comportemental** est utilisé ?\\

Exemples :\\
- <annotate>Cette information a beaucoup énervé Marie.</annotate> -> non.\\
- <annotate>Paul sanglote.</annotate> -> oui (car la tristesse de Paul est exprimée par le comportement "sanglote").\\
- <annotate>Avant d'arriver devant une salle de classe, les enseignants, eux aussi, sont sur les bancs de l'école.</annotate> -> non\\
- <annotate>Le père de Jeanne est mort hier à cause d'un cancer.</annotate> -> non.\\
- <annotate>Andrée a sautillé partout en chantant.</annotate> -> oui (car la joie de Andrée est exprimée par le comportement "sautillé partout en chantant").\\

Annotation (oui/non) :\\
- Nicolas Hulot n’appartient à aucun parti politique. <annotate>Il a refusé trois fois le poste de ministre de l’Ecologie avant d’accepter la proposition d’Emmanuel Macron.</annotate> Mais ça ne s’est pas très bien passé.  -> \\

\textbf{Assistant:}\\
\textit{réponse du modèle}\\

\textbf{User:}\\
Définition : Une émotion est dite du mode "montré" lorsqu'elle est exprimée par des caractéristiques linguistiques de l'énoncé qui traduisent l'état émotionnel dans lequel se trouvait l'énonciateur au moment de l'énonciation.\\

Question : Si la phrase à annoter est émotionnelle, est-ce que le mode **montré** est utilisé ?\\

Exemples :\\
- <annotate>Andrée a sautillé partout en chantant.</annotate> -> non.\\
- <annotate>Paul sanglote.</annotate> -> non.\\
- <annotate>Oh, chouette ! Quelle bonne idée !</annotate> -> oui (car la joie de l'énonciateur est traduite au sein de l'énoncé par les interjections "oh" et "chouette", les énoncés averbaux et les points d'exclamations).\\
- <annotate>Oh, non... C'est vraiment dommage !</annotate> -> oui (car la tristesse de l'énonciateur est traduite au sein de l'énoncé par l'interjection "oh", l'énoncé averbal, les points de suspension et le point d'exclamation.)\\
- <annotate>Avant d'arriver devant une salle de classe, les enseignants, eux aussi, sont sur les bancs de l'école.</annotate> -> non\\

Annotation (oui/non) :\\
- Nicolas Hulot n’appartient à aucun parti politique. <annotate>Il a refusé trois fois le poste de ministre de l’Ecologie avant d’accepter la proposition d’Emmanuel Macron.</annotate> Mais ça ne s’est pas très bien passé.  -> \\

\textbf{Assistant:}\\
\textit{réponse du modèle}\\

\textbf{User:}\\
Définition : Une émotion est dite "suggérée" lorsqu'elle est exprimée par la description d'une situation associée de manière conventionnelle à un ressenti émotionnel.\\

Question : Si la phrase à annoter est émotionnelle, est-ce que le mode **suggéré** est utilisé ?\\

Exemples :\\
- <annotate>Oh, chouette ! Quelle bonne idée !</annotate> -> non.\\
- <annotate>Le père de Jeanne est mort hier à cause d'un cancer.</annotate> -> oui (car où la tristesse de Jeanne est suggérée par la description du décès, il y a peu de temps, de son père, une personne proche d'elle).\\
- <annotate>Ces deux amis se retrouvent après une longue séparation.</annotate> -> oui (car la joie des deux amis est suggérée par la description de leurs retrouvailles après un temps long).\\
- <annotate>De 2007 à 2012, il a été le Premier ministre de l'ancien président Nicolas Sarkozy.</annotate> -> non\\
- <annotate>Andrée a sautillé partout en chantant.</annotate> -> non.\\

Annotation (oui/non) :\\
- Nicolas Hulot n’appartient à aucun parti politique. <annotate>Il a refusé trois fois le poste de ministre de l’Ecologie avant d’accepter la proposition d’Emmanuel Macron.</annotate> Mais ça ne s’est pas très bien passé.  -> \\

\textbf{Assistant:}\\
\textit{réponse du modèle}
}

% \section{Example Appendix}
% \label{sec:appendix}

% This is an appendix.

\end{document}